\documentclass[journal]{IEEEtran}  

\IEEEoverridecommandlockouts                              

\usepackage[ruled,vlined]{algorithm2e}

\usepackage{subfig, bm}

\usepackage{graphicx} 
\usepackage{amsmath} 
\usepackage{amsfonts}
\usepackage{amssymb}  

\title{\LARGE \bf
BiConMP: A Nonlinear Model Predictive Control Framework for Whole Body Motion Planning}

\author{Avadesh Meduri$^{*1}$, Paarth Shah$^{*3}$, Julian Viereck$^{1}$, Majid Khadiv$^{2}$, Ioannis Havoutis$^{3}$ and Ludovic Righetti$^{1,2}$
\thanks{$^{*}$Both the authors contributed equally.}
\thanks{This work was supported by New York University, the European Union's Horizon 2020 research and innovation program (grant agreement 780684) and the National Science Foundation (grants 1825993 and 1925079). Paarth Shah was supported by an AWS Lighthouse Scholarship. }
\thanks{$^{1}$Tandon School of Engineering, New York University (NYU), USA. 
        {\tt\small am9789@nyu.edu, ludovic.righetti@nyu.edu}}%
\thanks{$^{2}$Max-Planck Institute for Intelligent Systems, Tuebingen, Germany.
{\tt\small majid.khadiv@tuebingen.mpg.de}}%
\thanks{$^{3}$Oxford Robotics Institute, University of Oxford, England.
{\tt\small paarth@oxfordrobotics.institute, ioannis@oxfordrobotics.institute}}%
}

\usepackage{subfig}
\usepackage{hyperref}
\newtheorem{remark}{\textbf{Remark}}

\usepackage[dvipsnames]{xcolor}

\begin{document}

\maketitle
\thispagestyle{empty}
\pagestyle{empty}

\begin{abstract}
Online planning of whole-body motions for legged robots is challenging due to the inherent nonlinearity in the robot dynamics. In this work, we propose a nonlinear MPC framework, the BiConMP which can generate whole body trajectories online by efficiently exploiting the structure of the robot dynamics. BiConMP is used to generate various cyclic gaits on a real quadruped robot and its performance is evaluated on different terrain, countering unforeseen pushes and transitioning online between different gaits. Further, the ability of BiConMP to generate non-trivial acyclic whole-body dynamic motions on the robot is presented. The same approach is also used to generate various dynamic motions in MPC on a humanoid robot (Talos) and another quadruped robot (AnYmal) in simulation. Finally, an extensive empirical analysis on the effects of planning horizon and frequency on the nonlinear MPC framework is reported and discussed.   
\end{abstract}

\section{INTRODUCTION}
Legged robots can autonomously navigate and operate in environments built for humans. The efficacy of such robots depends largely on how efficiently they can move around, adapt to changes in their surroundings and recover from unforeseen disturbances. These decisions are usually made by trajectory optimization algorithms which compute optimal robot movements and contact forces and contact planners which decide which end-effector should make contact with the environment. Consequently, it is important for the trajectory optimization algorithm to be general enough such that it is able to generate any behaviour that is needed for the robot to achieve a desired task. At the same time, these algorithms should run as fast as possible such that they adapt online to any changes in the environment. In this work, we propose a general approach to generate whole body trajectories that is sufficiently fast to be used in a model predictive control (MPC) fashion.

Initially, algorithms based on simplified models such as the linear inverted pendulum model (LIPM) \cite{kajita2003biped, wieber2006trajectory} were developed to generate trajectories online for humanoid robots. These algorithms make use of a predefined footstep sequence provided by the user to generate a feasible center of mass (CoM) trajectory. Since the LIPM leads to an optimization problem with quadratic costs and linear constraints, the problem can be solved quickly using a quadratic program (QP) \cite{nocedal2006numerical}. While further extension of these algorithms enabled adaptation of the step location and timing \cite{herdt2010online,khadiv2016step, hopkins}, they are only capable of generating walking motions for flat grounds with co-planar contacts. 

On the other hand, frameworks that can plan contacts and optimal motions for complex scenarios have also been developed. In \cite{mordatch2012discovery,lengagne2013generation, posa2014direct}, the full-body motion and contact selection problems are formulated as single nonlinear optimization problems. In \cite{winkler2018gait}, a more efficient phase-based formulation of contact planning is proposed. Furthermore, \cite{tassasynthesis} makes use of differential dynamic programming (DDP) \cite{jacobson1968new} to solve the motion optimization problem through contact. While these approaches can in principle find complex contact sequences, they tend to be computationally too expensive to be used in real-time. Although \cite{neunert2018whole} showed nonlinear model predictive control on a quadruped using a Gauss-Newton multiple shooting variation of DDP and a relaxed spring damper contact model, results were limited to motions with low angular momentum such as trotting and jumping in place. 

Classically, the trajectory optimization problem has been split into two different sub-problems: contact planning and motion optimization. This decomposition reduces the complexity of the overall problem which allows them to be tractable. The main idea is to first generate a contact sequence given the terrain around the robot. The contact sequence is then provided to the motion planner to generate a feasible trajectory for the robot. The contact planning sub-problem can be solved using a variety of approaches such as mixed integer optimization \cite{deits2014footstep,ponton2016convex}, L1-loss based optimization \cite{tonneau2020sl1m},  graph search \cite{lin2019efficient} and sampling based approach \cite{tonneau2018efficient}. For the motion optimization sub-problem, the nonlinear dynamics can be split into two components, the actuated and unactuated dynamics (centroidal dynamics) \cite{wieber2006holonomy}. One of the interesting approaches to generate whole body motions quickly is to use the centroidal dynamics and full kinematics of the robot in one optimization problem \cite{dai2014whole}. Further,  a feasible whole-body motion can be generated more efficiently by iteratively optimizing for the centroidal dynamics and whole-body kinematics problems \cite{herzog2016structured}. Despite splitting the trajectory optimization problem into two parts, each individual sub-problem remains nonlinear and challenging to solve in real-time.

In order to further reduce computation times, several relaxations have been proposed to the centroidal dynamics formulation. In \cite{ponton2021efficient},  sequential convex relaxations are used and each relaxation is solved using a second order cone program \cite{alizadeh2003second}. Even though this approach can be used to quickly optimize a variety of motions, the reduction in compute times are not yet sufficient for closed-loop optimization. Further, in our experience, the relaxations are often not tight enough for very dynamic motions. Another approach to convexify the centroidal dynamics problem is to only minimize the worst case $L_{1}$ bound on the angular momentum \cite{dai2016planning}. While this formulation allows to solve the motion planning problem with a QP, it was only shown to be capable of generating motions with low angular momentum such as walking. For quadrupeds, with negligible leg inertia, the centroidal dynamics is also often approximated by linearizing the base rotation \cite{didynamic, repFree}. In these approaches, the swing foot trajectories are predefined which restrict the possibilities for whole-body motions and may lead to physical inconsistency. In \cite{sleiman2021unified}, DDP is used to solve an optimization problem with the centroidal dynamics and first order kinematics in real-time. This approach achieved re-planning frequencies suitable for real-time use. However, the approach relied on a low-level whole body controller and it is unclear whether this controller solely tracked the motions generated by DDP or whether the low-level controller behaved as a dynamic filter to ensure physical consistency \cite{yamane2003dynamics}. 
From our experience, the solve times for such methods increase for more dynamic motions such as bounding or rapid (forward) jumping due to the drastic changes in the momenta profiles.

In this work, we propose a nonlinear trajectory optimization framework that can be used in a real-time closed-loop model predictive control to generate whole-body motions using the kino-dynamic decomposition proposed in \cite{herzog2016structured}. The dynamics optimization problem is solved efficiently by exploiting the biconvex structure of the centroidal dynamics. We previously explored this structure in \cite{shah2021rapid} which leveraged the biconvex nature of the problem to formulate two separate, convex, sub-problems. Given the convexity of each sub-problem, an alternating procedure based on block coordinate descent was used which allowed the use of state of the art QP solvers and resulted in a speedup in solve times. However, very little is understood about the convergence rates of block coordinate descent which makes it unreliable for MPC where new solutions are needed in a fixed time\cite{shen2017disciplined}. In this work, we explore a different approach that also exploits the biconvex structure of the centroidal dynamics but we formulate the optimization problem using the Alternating Direction Method of Multipliers (ADMM) \cite{boyd2011distributed}, leading to a more efficient and reliable algorithm. We also split the biconvex dynamics differently which reduces the number of optimization variables.

Compared to the block coordinate descent algorithm, the ADMM algorithm provides favourable convergence properties such as the ability to reach acceptable solutions in fewer iterations and guaranteed sublinear convergence \cite{boyd2011distributed}. Crucially, due to the unconstrained nature of each sub-problem, each iteration is computationally cheap with respect to wall time which allows us to exploit the aforementioned convergence properties and make it attractive for use in an MPC fashion. In the proposed ADMM formulation, each convex sub-problem is solved using a custom implementation of the Fast Iterative Shrinkage Thresholding Algorithm (FISTA) \cite{beck2017first}. This approach guarantees quadratic convergence of the convex sub-problems while enforcing a variety of constraints including second order friction cone constraints. Our custom implementation exploits 1) the accelerated gradient nature of FISTA by warm starting the line search step to reduce solve times and 2) the sparse nature of our optimal control problem (OCP) as each iteration of the line-search only involves sparse matrix-vector multiplication which is less expensive compared to typical Quadratic Program (QP) or Quadratically constrained QP solvers which often involve expensive matrix decompositions. Due to this formulation, we solve the exact centroidal optimization problem without relaxing the friction cone constraints \cite{grandia2019feedback} or the dynamic constraints \cite{ponton2021efficient}. Finally, the first order optimization procedure of FISTA increases robustness to convergence in the absence of true gradients \cite{mohammadi2020robustness,liang2017activity}. This situation often occurs on real robots due to lack of accurate sensor measurements. Furthermore, problems such as ill-conditioning of the Hessians that second order methods like DDP encounter are also absent with FISTA.

We also propose a second order kinematics optimization formulation to generate smooth joint trajectories that track centeroidal momentum profiles as required in the kino-dynamic setup. We choose to solve this kinematics problem with a DDP solver \cite{mastalli2020crocoddyl} to exploit the sparsity in the problem. The second order nonlinear optimization removes the need to specify heuristic-based end-effector trajectories (e.g. via a spline based swing foot trajectory) as is often done in MPC implementations \cite{didynamic,khadiv2020walking}. Although these methods of swing-foot generation work well for simple motions, they are often restrictive in nature and do not allow the algorithm the freedom to find trajectories which may utilize the full capabilities of the end-effectors. The nonlinear kinematics solver generates non-trivial swing foot trajectories to track the desired centroidal momentum provided by the dynamics optimization. Further, the automatic generation of smooth joint acceleration profiles allows direct computation and tracking of torques on the robot with a simple inverse dynamics controller. This removes the need for a complicated QP-based whole-body controller that often behaves as an additional dynamic filter.

We demonstrate our approach in closed-loop MPC on the real Solo12 quadruped robot \cite{grimminger2020open} at 20 Hz to generate several gaits such as trotting, jumping and bounding. We also display the robustness of the framework against external disturbances and terrain noise. A high-five motion is also shown to demonstrate the generality of the approach to non-trivial, acyclic motions. Since the framework does not relax or impose assumptions to make the original problem convex, we are able to generate a wide array of motions in real-time. Furthermore, the same approach is used to generate various dynamic motions in MPC on a humanoid robot (Talos \cite{stasse2017talos}) and another quadruped robot (AnYmal \cite{hutter2016AnYmal}) in simulation. This underlines the generality,  robustness and low computation times of our proposed framework, BiConMP, despite the changes in the robot mass distributions, number of joints (size of optimization problem) and nature of the robot. This is often not as easy with other approaches which use simplified dynamics whose assumptions may not hold valid for different robots.  Finally, we empirically analyze the effects of the horizon length and re-planning frequency on the robustness and performance of the nonlinear MPC on the real robot (Solo12). To the best of our knowledge, this is the first reported empirical analysis of closed loop nonlinear MPC performance for legged robots. 

\begin{figure*}
\vspace{0.3cm}
\centering
\includegraphics[scale = 1.0]{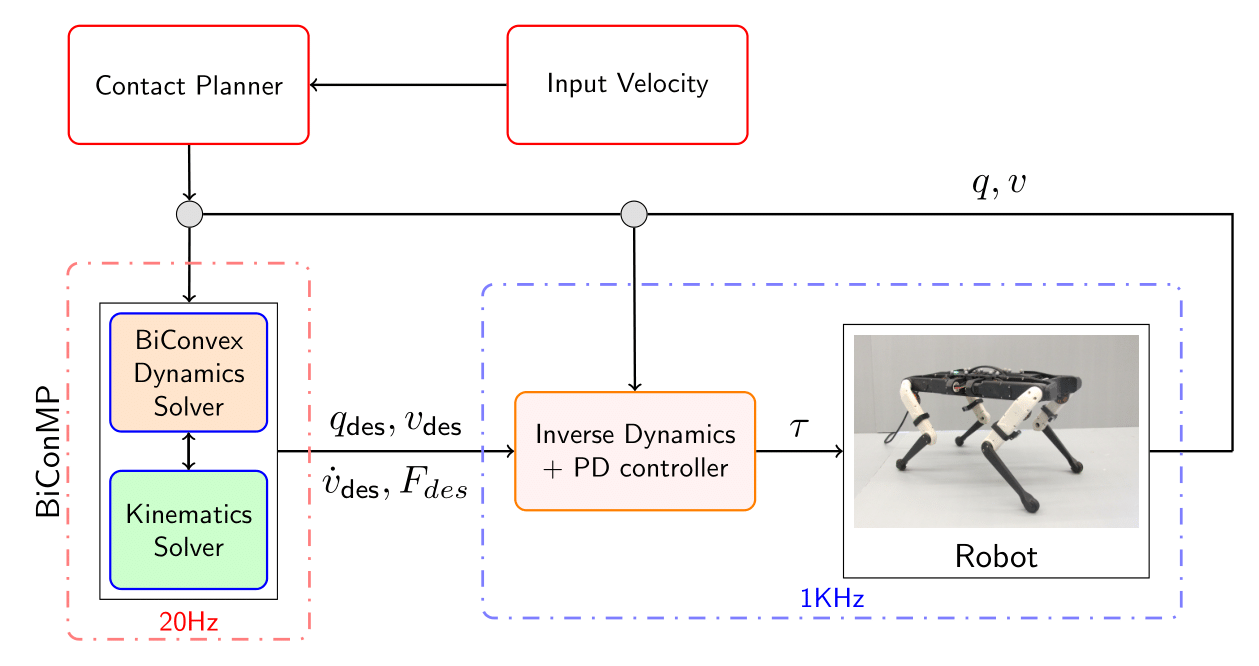}
\caption{A birds eye view of the entire nonlinear MPC framework. First, centroidal trajectories are generated using the ADMM framework explained in Section \ref{bicon}. These trajectories are used within a DDP-based kinematic optimizer that generates the desired joint trajectories (Section \ref{kin_optimizer}). The optimal force and joint trajectories from this kino-dynamic iteration are recomputed at 20 Hz and are used in an unconstrained inverse dynamics \eqref{rnea_inv_dyn} to compute the desired joint torques at 1 KHz. Finally these actuator torques are summed up with a fixed low gain impedance joint controller that result in the torques sent to the robot actuators \eqref{inverse_dynamics}.}
\label{fig:birds_eye}
\end{figure*}

\section{Background}
This section introduces the background necessary to describe our approach. First, we discuss the centroidal momentum dynamics of a floating base robot. Next, we explain the kino-dynamic trajectory optimization scheme used to generate feasible whole-body multi-contact motions for legged robots. Finally, we briefly introduce an optimization technique used in the solver. 

\subsection{Centroidal Dynamics}

The rigid body dynamics of floating base robot can be described as 
\begin{equation}\label{rbd}
    \mathbf{M}(\mathbf{q})\dot{\mathbf{v}} + \mathbf{N}(\mathbf{q}, \mathbf{v}) = \mathbf{S}^{T}\bm{\tau} + \sum_{j=1}^{N} \mathbf{J}_j^{T}\boldsymbol{\lambda}_j
\end{equation}
%
%
%
where $\mathbf{q} \in \mathbb{R}^n \times SE(3)$ is the generalized configuration of the robot (joint positions and base pose), and $\mathbf{v} \in \mathbb{R}^{n+6}$ is the generalized velocity vector. $\mathbf{M(q)} \in \mathbb{R}^{(n+6)\times(n+6)}$ is the mass matrix for the given robot configuration, $\mathbf{N(q, v)} \in \mathbb{R}^{n+6}$ is the vector containing all generalized forces (Coriolis, centrifugal, gravity, etc), $\bm{\tau} \in \mathbb{R}^n$ is the vector of joint torques, $\mathbf{S}$ is a selection matrix reflecting the underactuation of the robot, $\mathbf{J_{j}} \in \mathbb{R}^{6\times(n+6)}$ are the end effector Jacobians and $\boldsymbol{\lambda}_{j} = [\mathbf{f}_{j}, \bm{\kappa}_{j}] \in \mathbb{R}^6$ is the vector of forces and torques applied at each end effector.

The dynamics can further be split into its actuated and unactuated parts \cite{wieber2006holonomy, herzog2016structured}
\begin{subequations}\label{dynamics_split}
    \begin{align}
        \mathbf{M}_{u}(\mathbf{q})\dot{\mathbf{v}} + \mathbf{N}_{u}(\mathbf{q}, \mathbf{v}) &= \sum_{j=1}^{N} \mathbf{J}_{u,j}^{T} \boldsymbol{\lambda}_j \\
        \mathbf{M}_{a}(\mathbf{q})\dot{\mathbf{v}} + \mathbf{N}_{a}(\mathbf{q}, \mathbf{v}) &= \bm{\tau} + \sum_{j=1}^{N} \mathbf{J}_{a,j}^{T} \boldsymbol{\lambda}_j\label{eq:actuated_dynamics}
    \end{align}
\end{subequations}
where the subscript $a, u$ correspond to actuated and unactuated dynamics respectively. The unactuated dynamics is equivalent to the Newton-Euler equations of the center of mass (CoM)
\begin{equation}\label{newton_euler}
    \begin{bmatrix}\dot{\mathbf{l}} \\ \dot{\mathbf{k}}\end{bmatrix} = \begin{bmatrix} m \mathbf{g} + \sum_{j=1}^{N} n_{j}\mathbf{f}_{j} \\ 
    \sum_{j=1}^{N} n_{j}((\mathbf{r}_{j} - \mathbf{c}) \times \mathbf{f}_{j} + \bm{\kappa}_{j}) \end{bmatrix}
\end{equation}

where $\mathbf{l}, \mathbf{k}$ are the linear and angular momentum \cite{orin2013centroidal}, $m$ is the robot mass, $\mathbf{g}$ is the gravity vector, $\mathbf{c}$ represents the center of mass CoM location, $n_{j}$ is a binary integer that describes whether the end effector $j$ is in contact, $\mathbf{f}_{j}, \bm{\kappa}_{j}, \mathbf{r}_{j}$ are the end effector force, torque and location respectively. The linear momentum is related to the CoM velocity $\dot{\mathbf{c}}$  as $\mathbf{l} = m \dot{\mathbf{c}}$. The linear momentum and angular momentum can also be described in terms of the generalized joint configuration using the centroidal momentum matrix($\mathbf{D(q)}$) of the robot as $\begin{bmatrix}\mathbf{l} \\ \mathbf{k}\end{bmatrix} = \mathbf{D}(\mathbf{q})\mathbf{v}$, \cite{orin2013centroidal}. 


\subsection{Kino-Dynamic Motion Generation} \label{kino-dyn}
Splitting the dynamics enables multi-contact motion generation by only considering the unactuated dynamics or centroidal dynamics of the robot. Subsequently, a feasible whole-body trajectory can then be determined based on the centroidal plan and desired whole body tasks, provided there is sufficient torque authority \cite{herzog2016structured}\cite{ponton2021efficient}. This is an attractive approach since it breaks the original nonlinear optimization problem into two simpler sub-problems.

A desired motion plan using the centroidal dynamics can be generated by solving the following discrete optimal control problem (OCP)
\begin{align}
\underset{\mathbf{c}, \mathbf{\dot{c}}, \mathbf{k}, \mathbf{f}, \boldsymbol{\kappa}}{\text{min.}} & \quad \sum_{t = 0}^{T-1} \phi_{t}(\mathbf{c}_{t}, \mathbf{\dot {c}}_{t}, \mathbf{k}_{t}, \mathbf{f}_{t}, \boldsymbol{\kappa}_{t})
    + \phi_{T}(\mathbf{c}_{T}, \mathbf{\dot {c}}_{T}, \mathbf{k}_{T}, \mathbf{f}_{T}, \boldsymbol{\kappa}_{T}) \notag \\
    \text{s.t.} & \quad \mathbf{c}_{t+1} = \mathbf{c}_{t} + \mathbf{\dot{c}}_{t} \Delta t \label{pos_cst} \\
    & \quad \mathbf{\dot{c}}_{t+1} = \mathbf{\dot{c}}_{t} + \sum_{j=1}^{N} n^{j}_{t} \frac{\textbf{f}_{t}^{j}}{m} \Delta t + \mathbf{g}\Delta t \label{vel_cst} \\
    & \quad \mathbf{k}_{t+1} = \mathbf{k}_{t} + \sum_{j=1}^{N} n_{t}^{j}((\mathbf{r}^{j}_{t} - \mathbf{c}_{t}) \times \mathbf{f}^{j}_{t} + \boldsymbol{\kappa}^{j}_{t}) \Delta t \label{amom_cst} \\
    & \quad \forall_{t,j}, \: \sqrt{(\textbf{f}^{j}_{t, x})^{2} + (\textbf{f}^{j}_{t, y})^{2}} \leq \mu \textbf{f}^{j}_{t, z} \quad, 
    \textbf{f}^{j}_{t, z} \geq 0 \quad \\ 
    & \quad \forall_{t,j},\: \mathbf{r}^{j}_{t} \epsilon \Psi,  
    \quad \forall \mathbf{c}_{t} \in \Omega, 
    \quad \mathbf{c}_{0},\mathbf{\dot{c}}_{0} = \mathbf{c}_{\textit{init}}, \mathbf{\dot{c}}_{\textit{init}}
  \end{align}
where $\phi_{t}(\mathbf{c}_{t}, \dot{\mathbf{c}}_{t}, \mathbf{k}_{t}, \mathbf{f}_{t}, \boldsymbol{\kappa}_{t})$ is the running cost, $\phi_{T}(\mathbf{c}_{T}, \dot{\mathbf{c}}_{T}, \mathbf{k}_{T}, \mathbf{f}_{T}, \boldsymbol{\kappa}_{T})$ is the terminal cost, $\Delta t$ is the time discretization, $\mu$ is the friction coefficient, $\Psi$ is the set of all allowed stepping locations, $\Omega$ are kinematic constraints written as bounds on the CoM position, $\mathbf{c}_{\textit{init}}, \mathbf{\dot{c}}_{\textit{init}}$ are the initial conditions for the CoM.

The optimal joint trajectory is generated by solving a whole-body kinematic optimizer which tracks the optimal centroidal momentum obtained from the previous step using the centroidal momentum matrix, along with additional full body tasks, such as swing foot motion \cite{ponton2021efficient}. The generated momentum trajectory from the whole-body kinematic optimizer is then used as a soft constraint in the centroidal OCP to obtain refined centroidal and contact forces trajectories. This process is iterated until the two sub-problems converge \cite{herzog2016structured} leading to a solution to the original problem.

One can then directly use plain inverse dynamics to recover actuated joint torques from desired state trajectories and contact forces using Eq. \ref{eq:actuated_dynamics}
\begin{equation}\label{rnea_inv_dyn}
    \bm{\tau}_{RNEA} = \mathbf{M_{a}}(\mathbf{q)\dot{v}} + \mathbf{N}_{a}(\mathbf{q}, \mathbf{v}) - \sum_{j=1}^{N} \mathbf{J}_{a,j}^{T} \boldsymbol{\lambda}_j
\end{equation}

 Note that in this work we do not use a constrained QP-based inverse dynamics as is usually done, but we simply use the computed joint positions, velocities, accelerations and force trajectories in the Recursive Newton Euler Algorithm (RNEA) \cite{featherstone2014rigid} to compute the torques.

\subsection{Fast Iterative Shrinkage Thresolding Algorithm (FISTA)} \label{fista}

Proximal gradient methods \cite{beck2009fast} are a popular family of algorithms used to solve problems of the form  
\begin{equation}
\underset{\mathbf{x}}{\text{min.}} \quad T(\mathbf{x}) + I(\mathbf{x}) 
\end{equation}
where $\mathbf{x}$ is the optimization variable, $T(\mathbf{x})$ is the cost function to be optimized and $I(\mathbf{x})$ is usually an indicator function that enforces feasibility constraints or forces $\mathbf{x}$ to remain inside a feasible set. The cost function $T(\mathbf{x})$ can be non-smooth, nonlinear or convex and $I(\mathbf{x})$ is restricted to be convex. Each algorithm in the proximal gradient family varies slightly in the step length computation and update procedure for $\mathbf{x}$, however each iteration in the proximal methods is fundamentally of the form 
\begin{equation} \label{eq:prox}
 \mathbf{x}_{k+1} = P_{c}( \mathbf{x}_{k} + t_{k} \nabla f(\mathbf{x}_{k}))
\end{equation}
where $t_{k}$ is the step length, $\mathbf{x}_{k+1}$ is the value of the optimization variable at the next iteration ($k+1$) and $P_{c}$ is the proximal operator that ensures that after the descent step is taken, the new $\mathbf{x}_{k+1}$ lies within the domain of $I(\mathbf{x})$ \cite{beck2009fast}. Depending on the function represented by $I(\mathbf{x})$, the proximal operator may or may not have a closed form solution. The function $I(\mathbf{x})$ can only be used if it is possible to compute a closed form solution for the proximal operator \cite{beck2017first}. Consequently, arbitrary inequality constraints cannot be enforced with these methods. However, in the presence of closed form solutions (as is in our case) these proximal operators are very cheap to evaluate.  

In the specific case when the cost function $T(\mathbf{x})$ is convex, a proximal gradient method, the Fast Iterative Shrinkage Thresholding Algorithm (FISTA) is an attractive choice. FISTA is an accelerated first order gradient method that displays quadratic convergence. Algorithm \ref{FISTA} shows the steps in FISTA. The key point in the algorithm, as compared to other proximal methods, is the introduction of the auxiliary optimization variable $\mathbf{y}_{k}$ and the update procedure of $t_{k}$ which is the primary reason for the quadratic convergence. The step length $L_{k}$ is chosen based on a sufficient decrease condition (similar to Wolfe's condition \cite{nocedal2006numerical}). For more details regarding the algorithm we refer the reader to \cite{beck2009fast, beck2017first}.

\begin{algorithm}
\SetKwInOut{Input}{Input}
    \SetKwInOut{Output}{Output}
    Initialize optimization variables: $\mathbf{y}_{0} = \mathbf{x}_{0}, t_{0} = 1 $
    \\
    set $k = 0$ 
    \\
    \While{$k <\ $ {maximum iterations} } {
     Pick $L_{k} > 0$ \\
     $\mathbf{x}_{k+1} = prox_{\frac{1}{L_{k}}I}(\mathbf{y}_{k} + \frac{1}{L_{k}}\nabla f(\mathbf{y}_{k}))$ \\
     $t_{k+1} = \frac{1 + \sqrt{1 + 4t^{2}_{k}}}{2}$ \\
     $\mathbf{y}_{k+1} = \mathbf{x}_{k+1} + \frac{t_{k} - 1}{t_{k+1}}(\mathbf{x}_{k+1} - \mathbf{x}_{k})$
    }
    \caption{FISTA algorithm}
    \label{FISTA}
\end{algorithm}

In the following, we will exploit FISTA's quadratic convergence properties to quickly solve the convex sub-problems of the centroidal OCP (see Section \ref{conv_sub}). The indicator function in our formulation enforces kinematic (box constraints) and friction cone constraints (second order cone projections) for which the proximal operator exists. In practice, FISTA is computationally very cheap because it does not need the inversion of the Hessian to achieve quadratic convergence and the proximal step that enforces feasibility of $\mathbf{x}$ (inequality constraints) is inexpensive. 

\section{Approach}
We now introduce the main components of our solver, the BiConMP. First, we present the biconvex dynamics solver in detail and explain how it exploits the structure of the nonlinearity in the centroidal OCP to solve the problem efficiently. Second, we discuss the DDP based second order kinematics formulation used in the framework to solve the nonlinear problem. Finally, we give a birds eye view of how the BiConMP is used in a non-linear MPC setting to generate full body motions in real time. 

\subsection{Biconvexity in Centroidal Dynamics}
The unactuated dynamics constraints  \eqref{pos_cst}, \eqref{vel_cst}, \eqref{amom_cst} are nonlinear due to the cross product term in the angular momentum constraint \eqref{amom_cst}. This non-convexity makes the problem inherently difficult to solve. These constraints, however, have an interesting feature: they are biconvex \cite{boyd2011distributed}. That is, the constraints are affine in terms of $\mathbf{c}, \dot{\mathbf{c}}, \mathbf{k}$ when $\mathbf{f}_{t}^{j}, \forall_{t,j}$ is kept constant and vice-versa. Consequently, the terms of the discrete constraints \eqref{pos_cst}, \eqref{vel_cst}, \eqref{amom_cst} can be rearranged as an affine equation in terms of $\mathbf{X}$, $A(\mathbf{F})\mathbf{X} = b(\mathbf{F})$ and $\mathbf{F}$, $A(\mathbf{X})\mathbf{F} = b(\mathbf{X})$, where $\mathbf{X} = \{\mathbf{c}_{t}, \dot{\mathbf{c}}_{t}, \mathbf{k}_{t} \dots \}$ and $\mathbf{F} = \{\mathbf{f}^{j}_{t,x}, \mathbf{f}^{j}_{t,y}, \mathbf{f}^{j}_{t,z} \dots\}$, for $t = 0, \dots, T-1$, $j = 1, \dots, N$. Here $A(\mathbf{F})$ is a matrix whose elements depend on $\mathbf{F}$ and the centroidal dynamic constraints. Similarly, $A(\mathbf{X})$ is a matrix depending on $\mathbf{X}$ and dynamics constraints. $b(\mathbf{F})$ and $b(\mathbf{X})$ are vectors whose elements depend on $\mathbf{F}$ and $\mathbf{X}$ respectively. 

\subsection{Biconvex optimization with ADMM} \label{bicon}
Using the observation from the previous section, the centroidal dynamics OCP in Section \ref{kino-dyn} can be alternatively formulated as shown below, to highlight and exploit the biconvexity in the problem:
\begin{align}
    \underset{\mathbf{X} , \mathbf{F}}{\text{min.}} \quad &\Phi(\mathbf{X}) + I(\mathbf{X}) + \Phi(\mathbf{F}) + I(\mathbf{F}) \\
     \text{s.t.} \quad & \quad \quad  G(\mathbf{X},\mathbf{F}) = 0
\end{align}
where $\Phi(\mathbf{X}), \Phi(\mathbf{F})$ are the running and terminal cost functions in terms of $\mathbf{X}$ and $\mathbf{F}$ respectively, $G(\mathbf{X}, \mathbf{F})=0$ are the nonlinear constraints \eqref{pos_cst} such that it is bi-affine in terms of X and F, \eqref{vel_cst}, \eqref{amom_cst} and the initial state constraints ($\mathbf{c}_{0},\dot{\mathbf{c}}_{0} = \mathbf{c}_{\textit{init}}, \dot{\mathbf{c}}_{\textit{init}}$),  $I(\mathbf{X})$ is an indicator function that enforces kinematic constraints ($ \forall \mathbf{c}_{t} \in \Omega$, section \ref{kino-dyn}) while $I(\mathbf{F})$ is an indicator function that enforces unilaterality and friction cone constraints. 
This formulation makes it possible to exploit the biconvexity in the dynamics and solve the nonlinear problem very efficiently using the Alternating Direction Method of Multipliers (ADMM) \cite{boyd2011distributed}. We want to use the ADMM algorithm because it has an interesting property of reaching reasonably good solutions in a few iteration \cite{boyd2011distributed}. The downside however is that ADMM takes far more iterations as compared to a second order method (DDP) to obtain a high resolution result. In our application (closed loop MPC), solutions that have dynamic constraint violation tolerances lower than the sensor noise and satisfy  feasibility constraints are sufficient to be successfully deployed on the robot (as shown in our experiments). Consequently, when the need arises, ADMM allows us to terminate the solver before complete convergence (early termination) to ensure that the new trajectory is available in the desired time (real-time). At the same time, be sure that a reasonably good solution that is realizable on the robot is available. Furthermore, ADMM has a sub-linear convergence rates, which is not guaranteed with block-coordinate descent.

The proposed BiConMP solves the dynamics optimization by iteratively solving the two convex sub-problems (shown in Algorithm \ref{BiconvexMP}) as a part of a larger ADMM optimization scheme \cite{boyd2011distributed}. The ADMM algorithm solves both the convex sub-problems (force ($\mathbf{F}$) problem and state ($\mathbf{X}$) problem) iteratively until the dynamics violation falls below a desired tolerance (exit criteria). The dynamics violation is computed as
\begin{equation}
||A(\mathbf{F}_{k+1})\mathbf{X}_{k+1} - b(\mathbf{F}_{k+1})||^{2} \leq \epsilon_{dyn}
\end{equation}
where $\epsilon_{dyn}$ is the termination tolerance.

\begin{algorithm}
\SetKwInOut{Input}{Input}
    \SetKwInOut{Output}{Output}
    Initialize optimization variables: $F_{0}, X_{0}, P_{0}, \rho$
    \\
    set $k = 0$
    \\
    \While{$k <\ $ {maximum iterations} } {
     $\underset{F}{\text{min.}} \: \Phi(F) + \frac{\rho}{2}||A(X_{k})F - b(X_{k}) + P_{k}||^{2}+I(F)$ \\
     $\underset{X}{\text{min.}} \: \Phi(X) + \frac{\rho}{2}||A(F_{k+1})X-b(F_{k+1})+P_{k}||^{2}+I(X)$ \\
     $P_{k+1} = P_{k} +  A(F_{k+1})X_{k+1} - b(F_{k+1})$\\
      \If{$||A(F_{k+1})X_{k+1} - b(F_{k+1})||^{2} \leq \epsilon_{dyn}$}{
            terminate
        }
    }
    \caption{Biconvex Centroidal Optimization}
    \label{BiconvexMP}
\end{algorithm}

The cost function in the state sub-problem is always of the form $\Phi(\mathbf{X}) = (\mathbf{X} - \mathbf{X}_{nom})^{T}\mathbf{W}_{x}(\mathbf{X} - \mathbf{X}_{nom}) $
where $\mathbf{X}_{nom}$ is a nominal trajectory, $\mathbf{W}_{x}$ is a diagonal weight matrix.  The nominal trajectory gives the solver a heuristic idea of the desired centroidal trajectory (need not be dynamically consistent). In practice, the nominal trajectory usually consists of a desired base height and velocity (forward and sideways velocity). In addition, for some motions (cyclic gaits), a nominal angular momentum trajectory is provided to track a desired base orientation since direct orientation tracking is not possible with the centroidal OCP formulation (it is possible with the kinematics solver, section \ref{kin_optimizer}). The nominal angular momentum trajectory is computed  as follows:

\begin{equation}
    \mathbf{k}_{nom} = \mathbf{w}\: log_{3}(\mathbf{q}_{0} \ominus \mathbf{q}_{des})\label{eq:mom_orien_cost}
\end{equation}
 where $\mathbf{q}_{0}$ and $\mathbf{q}_{des}$ are the current and desired base orientation quaternions. $\ominus$ is the difference operator for quaternions and $log_{3}$ is the logarithmic map from $SE(3)$ to $\mathfrak{se}(3)$. $\mathbf{w}$ is a 3 dimensional weight vector. $\mathbf{k}_{nom}$ is set as the desired nominal angular momentum value for each time step in the planning horizon. In practice, the desired base orientation quaternion is always set to $[0,0,0,1]$ which corresponds to zero roll, pitch and yaw.
 
\begin{remark}
 Note that the exact Netwon-Euler dynamics are considered in the centroidal dynamics optimization. The inertia of the base is only ignored (assumes unit inertia) while computing the nominal angular momentum to be tracked in the cost (Eq. \eqref{eq:mom_orien_cost}). We use this heuristic computation because the resulting trajectories from BiConMP were able to track a desired orientation (even during external pushes) on the robot. A more accurate nominal trajectory could be used if the need arises. Note that the exact base orientation
 is optimized concurrently in the kinematic solver.
\end{remark}

In the force sub-problem, the cost is $\Phi(\mathbf{F}) = \mathbf{F}^{T}\mathbf{W}_{f}\mathbf{F}$ which penalizes unnecessary contact forces. To enforce complementarity constraints \cite{posa2014direct} based on the contact plan, coefficients of the variables corresponding to the time step for the given end effector where contact does not exist (elements of matrix $A(\mathbf{X}_{k})$), are set to zero in the force sub-problem. This automatically sets the planned forces to zero at that time step after optimization because of the cost function. 
\subsection{Convex Sub-problems}\label{conv_sub}
The state sub-problem can be solved using any Quadratic Program (QP). The force sub problem will need a Quadratically Constrained QP solver due to the second order friction cone constraints\cite{nocedal2006numerical}. These constraints could be relaxed and made linear, but this leads to more conservative motions (discussed later). In the BiConMP, we use FISTA (section \ref{fista}) because of the following favourable reasons: 1) FISTA maintains quadratic convergence even while enforcing constraints, 2) FISTA has low computation cost since its a first order method. The biconvex problem was formulated with indicator functions enforcing inequality constraints (kinematic and friction cone constraints) because FISTA can efficiently impose them using proximal operators, which are computationally inexpensive. 

\subsubsection{State sub-problem (optimizing for $\mathbf{X}$)} the kinematic constraints are enforced by the indicator function $I(\mathbf{X})$. This constrains the CoM to stay within a cube whose size depends on the location of the contact points at the particular time step. The proximal operator then becomes a box projection \cite{beck2017first} while computing a descent step in FISTA for the $k^{th}$ iteration, 
\begin{equation}
\mathbf{X}_{k+1} = \textit{max}(\textit{min}(\mathbf{X}_{k}^{*}, \mathbf{u}), \xi)
\end{equation}
where \[\mathbf{X}_{k}^{*} = \mathbf{Y}_{k} + \frac{1}{L_{k}}(\Phi^{'}(\mathbf{Y}_{k}) + \rho A(\mathbf{F}_{k})^{T}(A(\mathbf{F}_{k})\mathbf{Y}_{k} - b(\mathbf{F}_{k}) + \mathbf{P}_{k}))\] is the updated $\mathbf{X}$ parameter after the descent step is taken with $L_{k}$ as the line search step, $\mathbf{F}_{k}$ is an auxilary variable to $\mathbf{X}_{k}$ used in FISTA (subsection \ref{fista}), $\mathbf{u}$ and $\xi$ are the upper and lower bounds required to be satisfied for kinematic feasibility. For the components of $\mathbf{X}$ corresponding to velocity and angular momentum, the upper and lower bounds are set to $+\infty$ and $-\infty$ so as to enforce bound constraint only on the CoM location. 

\subsubsection{Force sub-problem (optimizing for $\mathbf{F}$)} the indicator function $I(\mathbf{F})$ enforces second order friction cone constraints \cite{bauschke1996projection}. The proximal operator enforcing this constraint for each group $f_{x}, f_{y},f_{z}$ corresponding to one contact point and time step in the $F$ vector is  
\[\begin{cases} 
      (0,0,0) & \mu \sqrt{(f_{k}^{x})^{2} + (f_{k}^{y})^{2}} \leq -f_{k}^{z} or f_{k}^{z} < 0  \\
      (\beta f_{x}, \beta f_{y}, \gamma f_{z}) & \mu \sqrt{(f_{k}^{x})^{2} + (f_{k}^{y})^{2}} > f_{k}^{z}  \\
      (f_{x}, f_{y}, f_{z}) & \sqrt{(f_{k}^{x})^{2} + (f_{k}^{y})^{2}} \leq \mu f_{z}
 \end{cases}
 \]
where $\mu$ is the friction coefficient,  
\begin{align}
\beta = \frac{\mu^{2}\sqrt{(f_{k}^{x})^{2} + (f_{k}^{y})^{2}} + \mu f_{z}}{(\mu^2 + 1)\sqrt{(f_{k}^{x})^{2} + (f_{k}^{y})^{2}}}
\end{align}
and 
\begin{align}
\gamma = \frac{\mu \sqrt{(f_{k}^{x})^{2} + (f_{k}^{y})^{2}} + f_{z}}{(\mu^2 + 1)}.
\end{align}
Subsequently, after a descent step is taken in FISTA to update the force vector 
\[\mathbf{F}_{k}^{*} = \mathbf{Y}_{k} + \frac{1}{L_{k}}(\Phi^{'}(\mathbf{Y}_{k}) + \rho A(\mathbf{X}_{k})^{T}(A(\mathbf{X}_{k})\mathbf{Y}_{k} - b(\mathbf{X}_{k}) + \mathbf{P}_{k}))\]
 every $f_{x}^{*}, f_{y}^{*},f_{z}^{*}$ in $\mathbf{F}^{*}_{k}$ is then projected based on the friction cone proximal operator to obtain the force vector $\mathbf{F}_{k+1}$ for the next iteration. Here $\mathbf{Y}_{k}$ is the auxiliary variable to $\mathbf{F}_{k}$ (section \ref{fista}). The projection of each group of forces independently works mathematically with the FISTA algorithm because the control decision variables are independent \cite{beck2017first} of each other in the centroidal problem. There is no explicit constraint enforcing unilaterality in $f_{z}$ because the friction cone projection implicitly enforce $f_{z} \geq 0$. The interesting point to note here is that with FISTA, the second order cone projection can be enforced directly while still maintaining quadratic convergence rates. In contrast, other QCQP solvers do not have quadratic convergence properties. They are also more computationally intensive because they need second order information. To improve solve time the friction cones are often approximated as linearized (i.e. converted to pyramidal polyedral constraints) in order to solve them with QPs. This usually results in conservative trajectory solutions which is not desirable when dynamic motions are to be performed.   

\subsubsection{FISTA implementation} To reduce the solve times in each iteration we specialize our implementation of the FISTA solver to exploit certain additional details specific to the dynamics optimization problem. Firstly, the analytical gradients of the cost function is used to compute the descent direction instead of using auto-diff or numerical differentiation methods. Secondly, the sparsity of the matrices $A(\mathbf{X}), A(\mathbf{F})$ are exploited during the matrix-matrix and matrix-vector computation in each iteration. Thirdly, the matrix multiplications, such as $A(\mathbf{X}_{k})^{T} A(\mathbf{X}_{k})$, $A(\mathbf{X}_{\mathbf{F}})^{T} A(\mathbf{X}_{\mathbf{F}})$, etc., which are only computed once in each convex sub-problem, are cached and reused. Finally, 
the accelerated gradient nature of the solver ensures that the first line search step is successful after certain number of iterations are reached. In practice, we noticed that almost the same step lengths were used in each iteration. Subsequently, we  warm start the solver with these line search steps which significantly improves the solve times as the solver no longer searches for the optimal values during run time.
The warm starting of the line search works well in FISTA because of an interesting property in its convergence proof which states that any line search parameter bigger than the Lipschitz constant of the  cost function will satisfy Wolfe's condition (chapter 10 \cite{beck2017first}). In this case, we conjecture that the empirically determined value satisfies this property. Warm starting the line search would not be possible conveniently with other QP solvers. 

\begin{remark}
The quadratic convergence property of FISTA, the computationally inexpensive proximal operators used to enforce the inequality constraints, and the above-mentioned details in the implementation significantly improve the solve times which play a crucial role in being able to re-plan online on the real robot. It is important to note that, thanks to FISTA, we have quadratic convergence for each sub-problem of the ADMM framework. We however only expect super-linear convergence for the whole centroidal trajectory optimization problem.
\end{remark}

\subsection{Kinematics Solver}\label{kin_optimizer}
The full-body kinematics trajectory generation problem described in section \ref{kino-dyn} is also nonlinear in nature. In the BiConMP, the problem is solved quickly using Differential Dynamic Programming (DDP) \cite{jacobson1968new}. DDP exploits the block diagonal structure  of the matrices while optimizing the problem and also shows quadratic convergence \cite{de1988differential, murray1984differential}. We use Crocoddyl \cite{mastalli2020crocoddyl}, an open-source DDP implementation. 
We formulate the problem as
\begin{align}
        \underset{\mathbf{q}, \mathbf{v}, \dot{\mathbf{v}}}{\text{min.}} \quad &\sum_{t = 0}^{T} \Phi_{mom}^{t}(\mathbf{l}^{*}_{t}, \mathbf{k}^{*}_{t}) + \Phi_{CoM}^{t}(\mathbf{c}^{*}_{t}) + \Phi^{t}_{eff}(\mathbf{q}_{t}, \mathbf{v}_{t}) + ||\dot{\mathbf{v}}|| \notag \\ 
        \text{s.t.} & \quad  \mathbf{q}_{t+1} = \mathbf{q}_{t} + \mathbf{v}_{t} \: \Delta t,
        \quad \mathbf{v}_{t+1} = \mathbf{v}_{t} + \dot{\mathbf{v}}_{t} \: \Delta t
\end{align}
where, $\Phi_{mom}^{t}(\mathbf{l}^{*}_{t}, \mathbf{k}^{*}_{t})= ||D(\mathbf{q}_{t})\mathbf{v}_{t} - \begin{bmatrix} \mathbf{l}^{*}_{t} \\ \mathbf{k}^{*}_{t}\end{bmatrix}||$ is a momentum cost that tracks the optimal linear and angular momentum computed by the centroidal OCP (Algorithm \ref{BiconvexMP}), $\Phi_{CoM}^{t}(\mathbf{c}^{*}_{t})$ is the center of mass tracking cost with the optimal CoM trajectory ($\mathbf{c}^{*}_{t}$) obtained from the centroidal OCP, $\Phi^{t}_{eff}(\mathbf{q}_{t}, \mathbf{v}_{t})$ is the end effector locations and velocity cost, and $||\dot{\mathbf{v}}||$ is a penalty on the control. In practice, the cost on the control encourages smooth motions of the end effectors especially during contact transitions. For example, during the landing phase of a jumping motion the kinematics solver retracts the legs of the robot in the air so that a large torque is not needed at the time of contact to bring the legs to rest (satisfy complementarity constraints). In essence, this reduces the impact of the legs during landing and makes the motion smooth on the real robot. This is one of the advantages of using a nonlinear kinematics formulation that plans a full body motion by taking the future into account. 

\subsection{The Model Predictive Control Pipeline}

An overview of the entire framework is shown in Fig. \ref{fig:birds_eye}. Given the current states of the robot $q_{init}, v_{init}, \dot{v}_{init}$, desired gait, planning horizon and velocity, a contact plan is either generated and adapted using the Raibert controller \cite{kim2019highly} or pre-defined without contact adaptation for acyclic or general motions. The BiConMP framework takes the input states and computes the optimal end effector forces, joint positions, joint velocities and joint acceleration trajectories for the entire horizon. Given the desired joint trajectories and contact forces, we use \eqref{rnea_inv_dyn} along with a low joint impedance around the desired states to compute the desired torques at 1 KHz (Equation \eqref{inverse_dynamics}). The desired torques is then sent to the robot which are tracked on board at 10 KHz. The BiConMP control loop is re-computed at 20 Hz (50 ms) to update for optimal motion and control trajectories in real time. 

In our BiConMP framework, we re-plan the whole-body trajectories every 50 ms, for a horizon larger then 50 ms, based on feedback from the current state of the robot. The feedforward torques are computed every 1 ms with inverse dynamics (Equation \eqref{rnea_inv_dyn}) based on the open loop trajectories in between two re-planning instances. Finally, a low joint impedance around the desired states are added to the computed torque to result in the final joint torques
\begin{equation}\label{inverse_dynamics}
    \mathbf{\tau}_{i} = \mathbf{\tau}_{RNEA,i} + K_{p}(\mathbf{q}_{d,i}-\mathbf{q}_{r,i}) + K_{y}(\mathbf{v}_{d,i} - \mathbf{v}_{r,i})
\end{equation}
where $\mathbf{\tau}_{i}$ is the torque sent to joint $i$, $\mathbf{\tau}_{RNEA,i}$ is computed using interpolated values of $\mathbf{f}_{i}$, $\mathbf{q}_{i}, \mathbf{v}_{i},$ and $\dot{\mathbf{v}}_{i}$ between each MPC cycle, $K_{p}$ and $K_{y}$ are the joint position and velocity impedance gains respectively. Subscripts $d$ and $r$ stand for desired and actual respectively. In the rest of the paper, we refer to the RNEA based controller \eqref{inverse_dynamics} as the inverse dynamics (ID) controller. 

When the provided contact plan is longer than the desired MPC horizon (e.g. for acyclic motions below), the plan is segmented into a smaller section matching the desired horizon length and then provided to the BiConMP. As time elapses, the segment is shifted (moving horizon) to select the part of the contact plan starting from the elapsed time $t$ and ending at $t+T$ where $T$ is the desired horizon. For gaited motions, the contact plan is automatically updated based on the time that has elapsed which determines which phase the legs should be in and for how long depending on the gait parameters and desired horizon length (horizon is kept constant after the start of the motion). The desired velocity is also updated at every cycle based on the user input. 

\begin{remark}
We would like to emphasize the importance of each individual component in our BiConMP pipeline. First, we exploit the biconvex structure in the centroidal dynamics and efficiently generate centroidal trajectories while respecting force constraints (Section \ref{bicon}). Second, for solving the convex sub-problems in the centroidal problem, we use FISTA which ensures quadratic convergence even with second order friction cone constraints. Third, we solve a simple second order whole-body kinematic optimizer given the momentum profiles provided by the centroidal dynamics optimization. This allows to drastically reduce computation time compared to the approaches using full-body dynamics DDP for the kinematics problem \cite{budhiraja2019dynamics}. Fourth, given the planned forces from centroidal MPC and desired joint trajectories from the IK, we compute the joint torques using Eq. \eqref{rnea_inv_dyn} without a need to solve a constrained whole-body inverse dynamics. Note that contrary to \cite{sleiman2021unified}, we intentionally did not use a constrained inverse dynamics on top of our MPC block to demonstrate the quality of the plans generated by our approach without the need to further filter them for physical consistency \cite{yamane2003dynamics}. Specifically, torque references computed from our MPC algorithm (which enforces friction cone constraints) can be applied directly to the robot.
\end{remark}

\section{Experiments}
In this section we present results obtained on a real Solo12 quadruped  \cite{grimminger2020open} along with simulation results on a humanoid and another quadruped robot. We first study the optimizer, including effects on the solve times as the size of optimization problem changes and the behaviour of termination criteria used in the biconvex dynamics solver. Next, we present the different motions (cyclic and acyclic) generated on Solo12 along with the performance of the BiConMP in various scenarios to test the robustness of our approach. Finally, we present simulation results on the AnYmal quadruped and the Talos humanoid to show that the approach can be directly applied to legged robot with other morphologies and mass distributions. The attached video illustrates all these experiments.

\subsection{Implementation details}
The entire BiConMP is implemented in C++. The biconvex dynamics optimizer is implemented from scratch including a custom implementation of FISTA. Croccoddyl \cite{mastalli2020crocoddyl} is used to solve the kinematics problem. The code will be made available open-source upon acceptance of this paper.  

All experiments were run on a Dell  precision  5820 tower  machine  with  a  3.7  GHz  Intel  Xeon  processor and rt-preempt kernel. Robot Operating System 2 (ROS 2) was used to handle the multi threading requirement of the approach to communicate between the 1 kHz inverse dynamics control loop and the 20 Hz MPC loop. The BiConMP is run on a node using a service client setup while the main inverse dynamics control loop (\ref{inverse_dynamics}) is run on a separate node at 1 kHz. The BiConMP service node is called at 20 Hz to update the plan and provide it to the inverse dynamics controller. The desired torque commands are then computed and provided to Solo12 via Ethernet channel.

\begin{figure*}
\vspace{0.3cm}
\centering
\includegraphics[width=17 cm, height=6.0 cm]{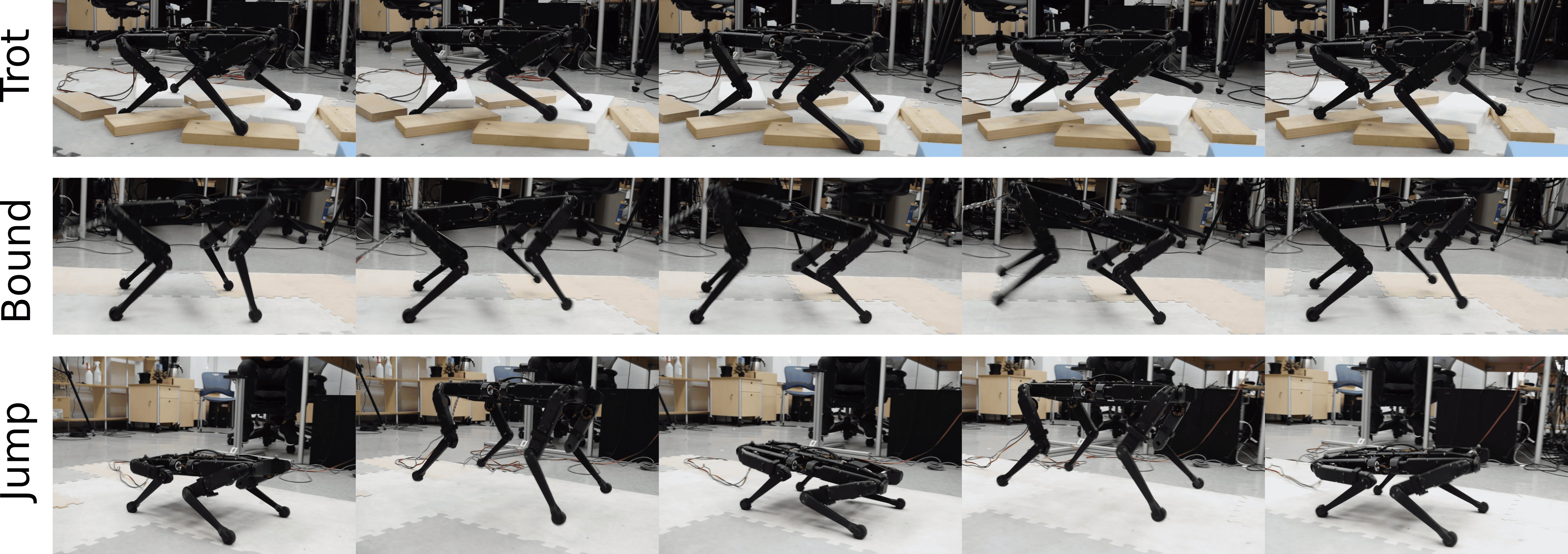}
\caption{Different motions demonstrated on real robot and simulation with solo12.}
\label{fig:motions}
\end{figure*}

\subsection{Solver Analysis}

\subsubsection{Solve Times} 
\begin{figure}
 \vspace{0.3cm}
    \centering
    \includegraphics[scale = 0.3]{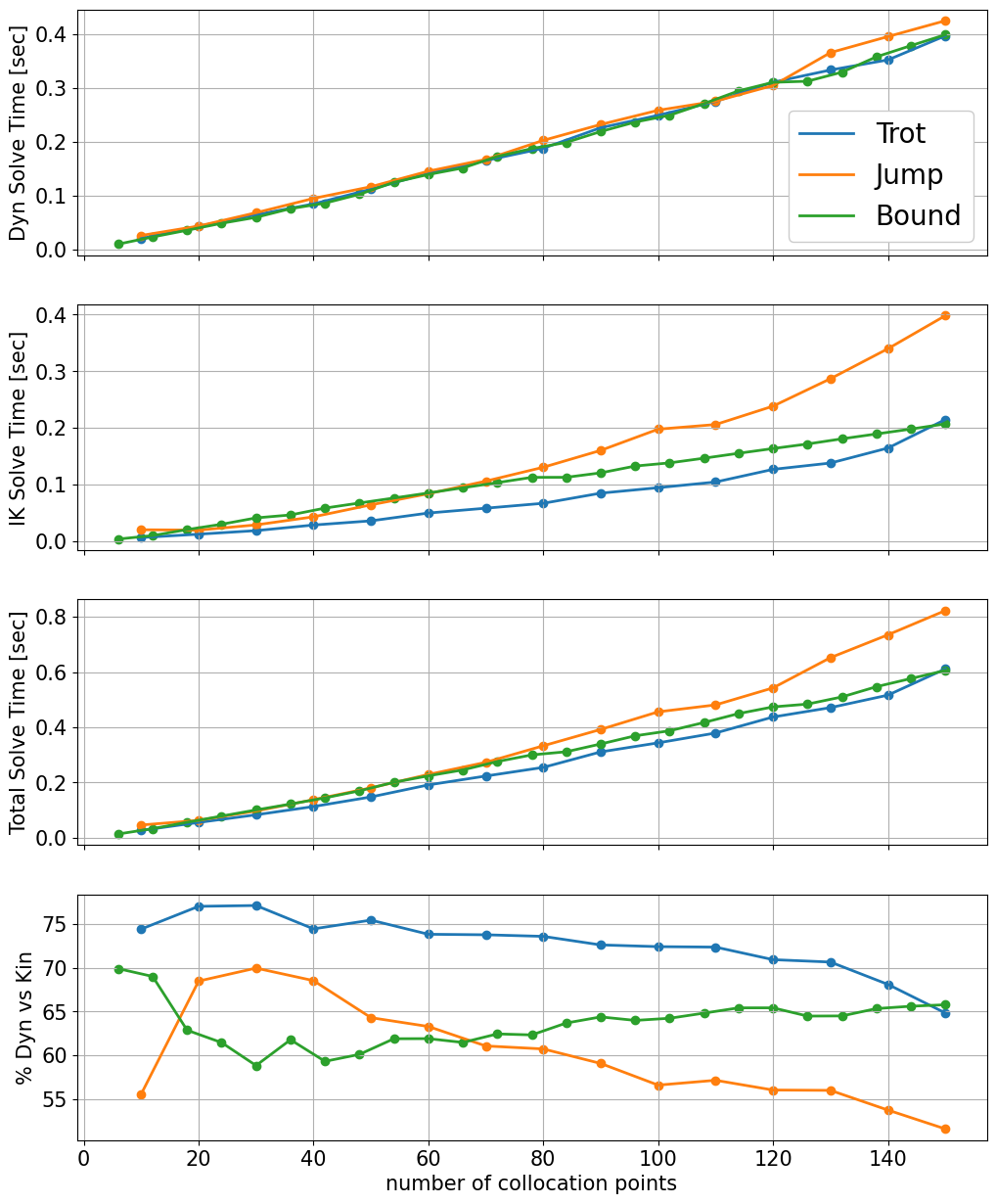}
    \caption{Solve times vs number of collocation points.}
    \vspace{-0.6cm}
    \label{fig:solver_vs_col}
\end{figure}

To analyze the solve times of the BiConMP as a function of the number of collocation points/problem size, three motions (trot, jump and bound) are used. These cyclic gaits are used as it is straightforward to change the horizon of the problem for this analysis. For each of these motions, the weights of the optimization problem, step time, discretization time, tolerances, etc., are kept the same and only the horizon of the problem is increased. In all the cases, the solver is terminated only after satisfying the termination criteria. The resulting solve times are shown in Fig. \ref{fig:solver_vs_col}. The top 3 plots contain the solve times from the dynamics biconvex solver, the DDP based kinematics solver and the total solve time (including miscellaneous operations like cost creations), respectively. The biconvex dynamics solver shows a linear increase in the solve times as the problem size increases. The kinematic solver also show an almost linear growth with increase in the number of collocation points. The solver does violate this trend at times depending on the termination criteria used in Crocoddyl \cite{mastalli2020crocoddyl}. However, the kinematics solver maintains a strong linear behavior in the problem size that is mainly important in this work to achieve MPC (between 6-12 collocation points).  On the other hand, the total solve times of the BiConMP framework maintains a linear growth in the solve rates. The solver always remains real-time, that is, it converges faster than the horizon of the plan. For example, the solver takes less than 0.85 seconds to generate a jump motion with a horizon of 7 seconds (140 collocation points $\times$ 0.05). The third plot also shows that the biconvex solver takes the most time in the framework. Consequently, further decrease in solve times can be achieved by warm starting the solver with pre-computed solutions or by highly optimizing the code \cite{lembono2020learning}.

\subsubsection{Termination Criteria} \label{termination}

\begin{figure}
 \vspace{0.3cm}
    \centering
    \includegraphics[scale = 0.5]{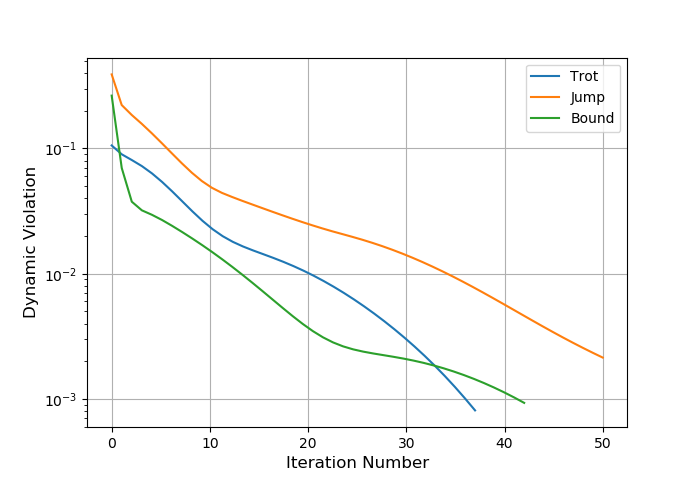}
    \caption{Dynamic violation as compared to iterations in the dynamics optimizer.}
    \vspace{-0.6cm}
    \label{fig:violation_vs}
\end{figure}

The dynamics violation is used as the termination criteria for the dynamics biconvex solver (section \ref{bicon}). The solver is terminated when the centroidal dynamics constraints fall below the threshold of 0.001 or until we hit a maximum number of ADMM iterations. Each iteration here refers to one ADMM iteration. In Fig. \ref{fig:violation_vs}, the dynamic violation vs the number of iterations is plotted for the three motions discussed previously. Each convex sub-problem in the biconvex ADMM problem is solved until the norm of the gradient falls below $1e-5$. A high tolerance is necessary to ensure that the main biconvex problem converges to high quality solutions. The dynamics violation rapidly decreases to a small value across all motions in alignment to the sublinear convergence property of the ADMM algorithm \cite{boyd2011distributed}. Note that a reasonable solution is found in a few iterations as reflected by the rapid drop in the dynamic violation. This is in alignment with the property of ADMM to get to roughly good solutions in a few iterations (section \ref{bicon}). This allows us to set a maximum number of iterations of the ADMM algorithm in order to guarantee the real-time performance of our framework. We found that in practice there are times where we may hit the maximum number of iterations before our solver finds a solution that satisfy our criteria for acceptable dynamic violation. Even though our solutions don't fall within our predefined threshold for dynamic violation, we found empirircally that these trajectories are of sufficient quality to be executed on the robot. 
In practice, setting the maximum number of ADMM iterations to 50 allowed us to run all of the required motions on the robot.
The kinematics solver is terminated based on the default settings provided by Crocoddyl. 


\subsection{Cyclic Gaits} \label{cyclic}
\begin{figure}
 \vspace{0.3cm}
    \centering
    \includegraphics[scale = 0.35]{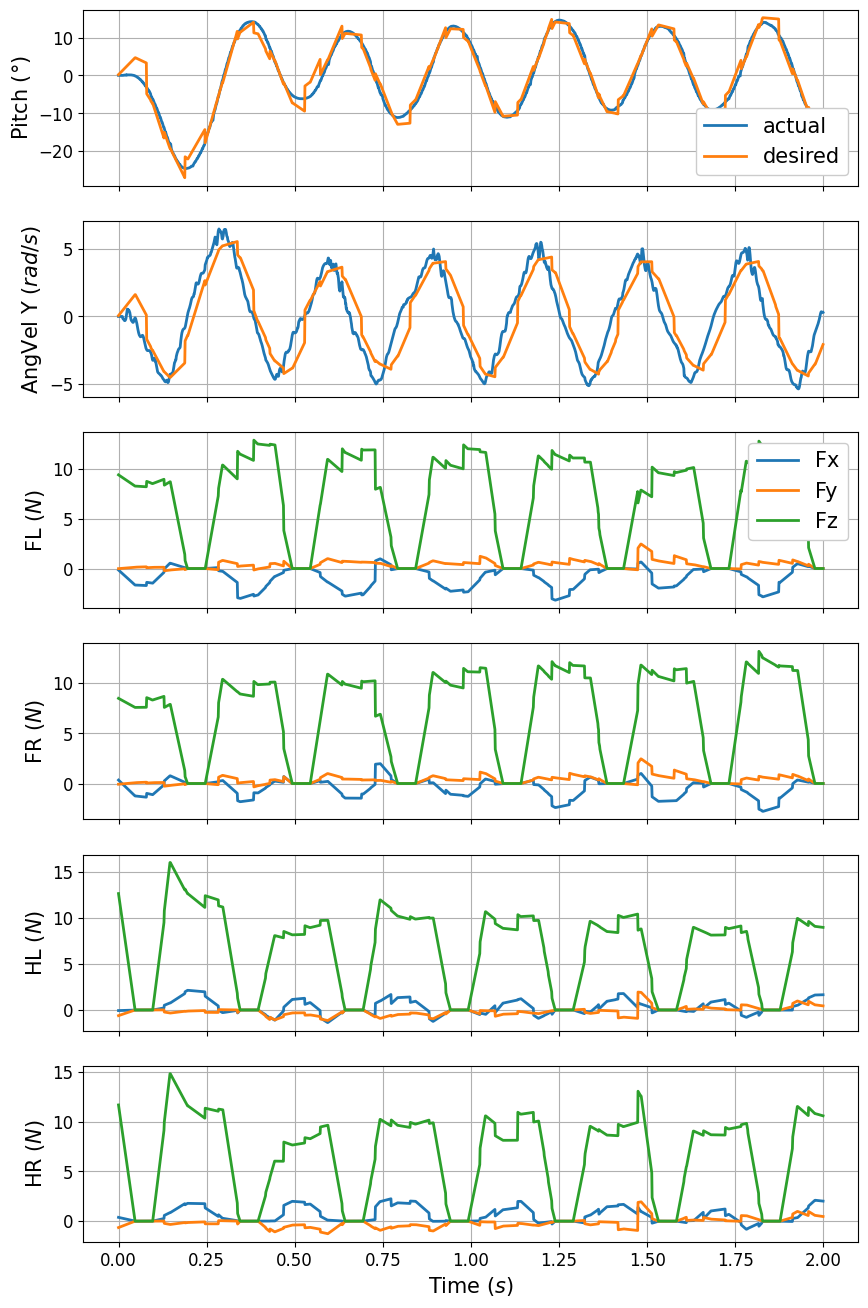}
    \caption{Angular velocity and forces during bounding gait.}
    \vspace{-0.6cm}
    \label{fig:amom_force_bound}
\end{figure}

We generated different gaits such as trots, jumps and bounds for the Solo12 quadruped.  The resulting motions are shown in Fig. \ref{fig:motions}. Table \ref{tab:gait_params} outlines the parameters used to design the gaits. 
Figure \ref{fig:amom_force_bound} shows the actual and desired base angular velocity about the Y axis (pitching axis) and the desired forces from the planner. The BiconMP is able to generate a bounding motion with considerable change in angular momentum and pitch magnitude.

\begin{table*}[h!]
\centering
\begin{tabular}{||c|c|c|c|c||}
\hline
Motion & Stance Duration (s) & Gait Duration (s) & Collocation Discretization $\Delta t$ (s) & Number Of Collocation Points \\
\hline
Trot   &     0.15        &     0.3      &    0.03           &   10   \\
\hline
Jump   &      0.2       &      0.5     &    0.05                       & 10   \\
\hline
Bound  &    0.15         &      0.3     &  0.05 & 12 \\
\hline
\end{tabular}
\caption{\label{tab:gait_params} Gait parameters for the various gaits tested on hardware.}
\end{table*}

We observed empirically that a smaller gait horizon than the ones chosen for each motion discussed above often led to the solver diverging after the completion of a few gait cycles. We hypothesize that this instability is due to the lack of a terminal cost/constraint that ensures the viability of the gait \cite{wieber2016modeling}. In the presence of a suitable terminal cost the horizon could be reduced \cite{mayne2014model}. A planning horizon of 2 gait cycles was necessary to ensure stability of the solver for bounding motions. The need for a longer planning horizon became especially necessary at higher speeds as these motions require tighter regulation of the angular momentum. Specifically, in order to create ground reaction forces to control motions with higher angular momentum, we believe a higher amount of control authority is required over time to bring the motion to a viable state that can be tracked successfully.

\begin{figure}
 \vspace{0.3cm}
    \centering
    \includegraphics[scale = 0.35]{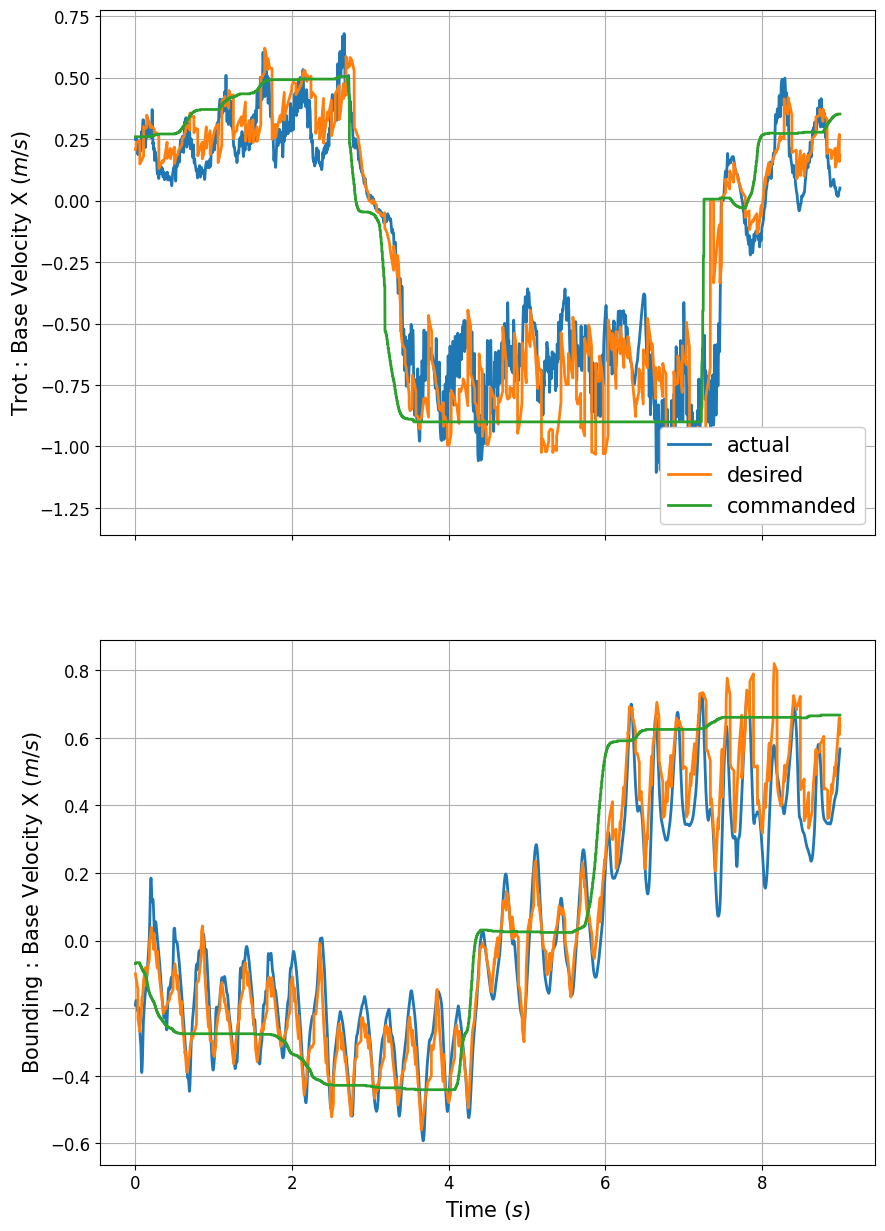}
    \caption{Velocity tracking for trot and bound gait on the real robot. x axis is the elapsed time in seconds. }
    \vspace{-0.6cm}
    \label{fig:bound_vel_track}
\end{figure}

The BiConMP is able to track desired linear and angular velocities accurately on the robot irrespective of the gait. In Fig. \ref{fig:bound_vel_track}, the velocity tracking performance for the trot and bounding gaits are shown. The framework is able to trade off the desired velocity input commands while staying within the limit cycle in real-time. The motions are also robust to unforeseen disturbances (push recovery) and unaccounted uneven terrain as shown in the accompanying video.

\begin{table*}[h!]
\centering
\begin{tabular}{||c|c|c|c|c|c||}
\hline
Gait & Scenario & Replan Frequency (Hz) & Mean Solve Time (ms) & Standard Deviation (ms) & Max Solve Time (ms)\\
\hline
Trot & Flat Ground & 20 & 23.47 &  2.45 & 32.8   \\
\hline
Trot & Push & 20 & 21.49 & 3.1 & 33.03 \\
\hline
Trot & Terrain & 20 &  26.13 & 4.45 & 36.5 \\
\hline
Jump & Flat Ground & 20 & 23.32 &  6.7 & 44.2   \\
\hline
Jump & Push & 20 & 22.45 & 3.4 & 37.8 \\
\hline
Jump & Terrain & 20 &  29.55 & 7.37 & 40.4 \\
\hline
Bound & Flat Ground & 20 & 27.15 &  5.2 & 42.6 \\
\hline
Bound & Push & 20 & 27.16 & 5.3 & 43.1 \\
\hline
Bound & Terrain & 20 &  27.27 & 5.3 & 40.0 \\
\hline
\end{tabular}
\caption{\label{tab:solve_times} Solve times of the BiConMP on Solo12 for various gaits and scenarios.}
\end{table*}

\subsubsection{Solve times in real experiments}
To evaluate the solve times, each gait (trot, jump and bound) are run on the robot in 3 different scenarios for $15$ seconds (300 replan calls of BiConMP), i) flat ground with no disturbance, ii) flat ground with external disturbances, iii) uneven terrain (step height of 5-8 cm ($20 - 32 \%$ of the nominal base height)) without external disturbances. Table. \ref{tab:solve_times} shows the solve time statistics for each of these scenarios. As can be seen, the max solve time (worst case solve time) does not exceed the replanning time of $50$ ms ($20$ Hz) and remains real-time regardless of the circumstances. Also, it is interesting to note that the mean solve times for each motion in the presence of terrain and pushes are quite similar to the flat ground scenario, which means that the solver remains unaffected by uncertain situations. On the other hand, the solve times with the same gait parameters / weights were lower on an average by about 5-6 milliseconds without sensor noise (such as in simulation). Hence, to achieve higher re-planning frequencies, obtaining clean sensor data seems important. 

The solve times of the framework for different motions range between 20-35 milliseconds on average, which is about 20-35 control cycles. Since the solve times are not negligible there exists a plan lag or delay from when sensor input is received and when the new plan is available. This lag has shown to cause instabilities on real robots during run times and several approaches have been proposed to deal with this issue \cite{koenemann2015whole}. In our experiments, we skip the section of the plan that falls between the planning time and track the rest of the plan with the inverse dynamics controller. That is, if a new plan is requested at $T = 0$ and the plan is available after $t$ milliseconds, the plan from $t$ milliseconds to the end is used assuming that the robot is close to the plan at $t$ milliseconds. Even though this might not necessarily be true all the time (for example - push recovery, terrain noise) this strategy did not affect the stability of the gaits on the robot when compared to simulation where time can be frozen until a new plan is available. 

\begin{figure}
 \vspace{0.3cm}
    \centering
    \includegraphics[scale = 0.35]{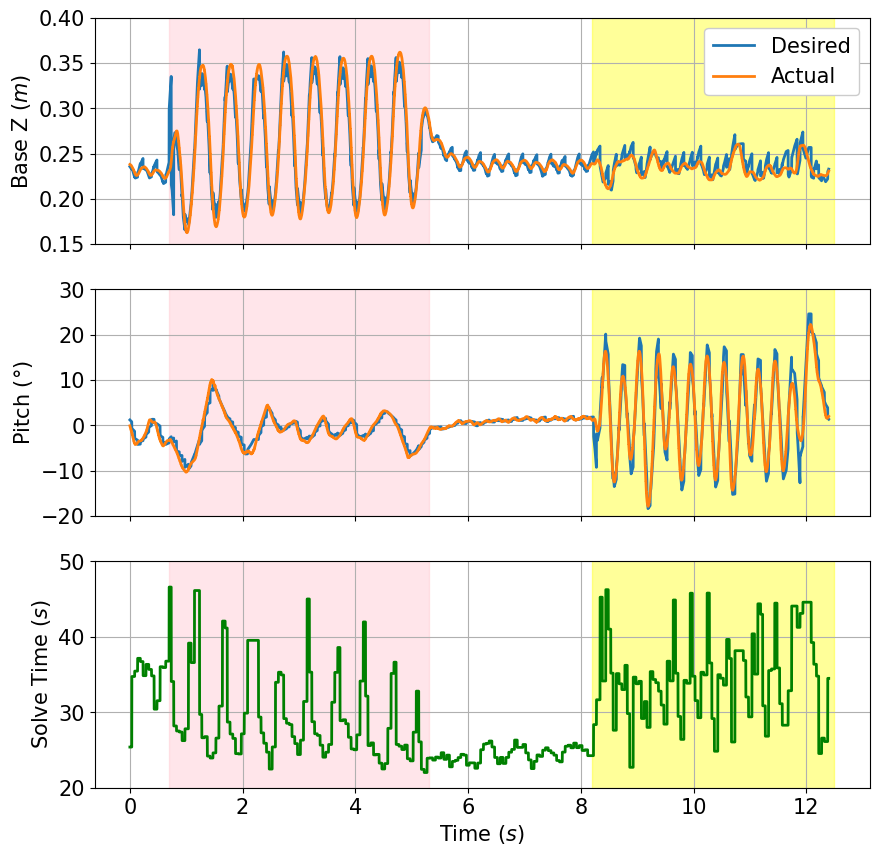}
    \caption{Solve times during live gait transitions. The white color regions corresponds to the trot gait, the pink section represents the jumping gait and the yellow section depicts the bounding phase.}
    \vspace{-0.6cm}
    \label{fig:gait_transition}
\end{figure} 

\subsubsection{Gait Transitions} Taking advantage of the re-planning capability of the framework, unplanned stable transitions are possible between the 3 different gaits. A plot of the base height and pitch of the base along with the solve times are shown in Fig. \ref{fig:gait_transition}. The robot is initially in a brief trot phase depicted by the almost constant pitch and base height. After which, the robot transitions to the jumping gait at around 1 second, as can be seen from the large amplitude oscillations in the base height. After about 5 seconds the robot transitions back to the trot gait for about 2 seconds. Finally, the robot moves to the bounding gait between 8 to 12 seconds as can be observed with the large changes in the pitch of the robot. All these transitions happen when the user desires these changes without any pre-planning between gaits changes. When the gait change is triggered, the gait parameters (Table \ref{tab:gait_params}) and contact plan (discussed in the previous subsection) is changed to the new gait. The BiConMP then generates a stable motion to transition to the desired gait. During all these transitions the framework keeps its solve times below 50 milliseconds. Also, the solve times for each gait remain similar to when there are no transitions (Table. \ref{tab:solve_times}).

\subsection{Acylic Motions} \label{acyclic}
\begin{figure*}
\vspace{0.3cm}
\centering
\includegraphics[width=17 cm, height=6.0 cm]{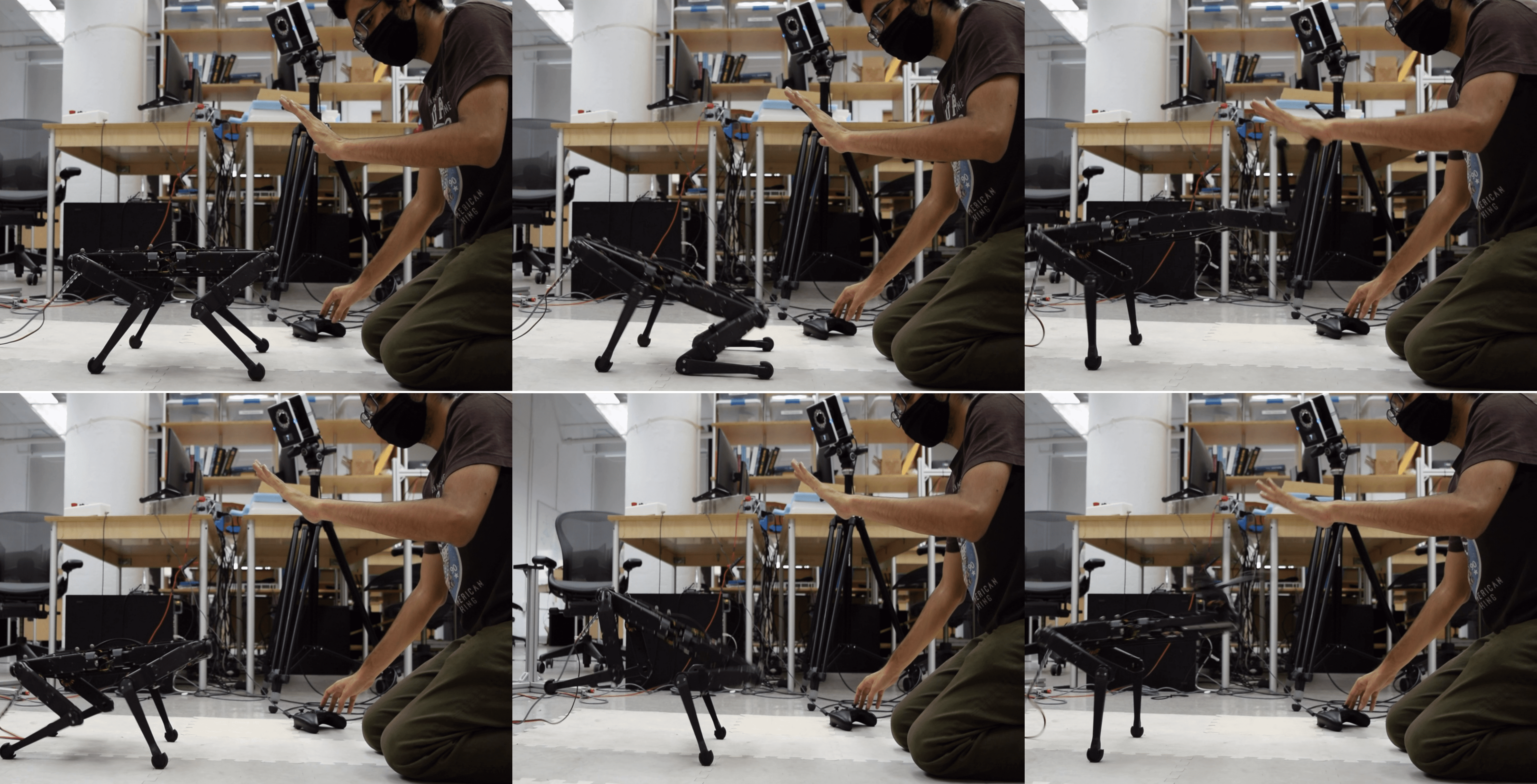}
\caption{High five motion (top left to bottom left in clock wise direction).}
\label{fig:hifive}
\vspace{-0.5cm}
\end{figure*}

\begin{figure}
 \vspace{0.3cm}
    \centering
    \includegraphics[scale = 0.5]{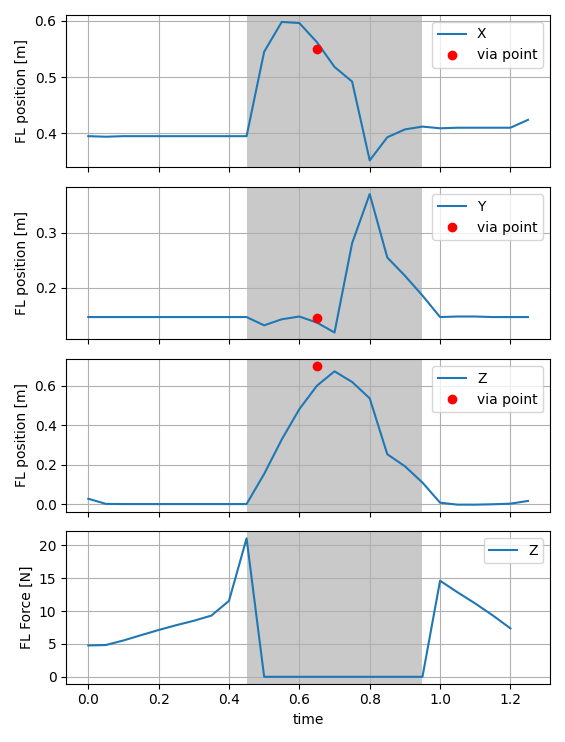}
    \caption{Swing foot trajectory of front left leg from the IK.}
    \vspace{-0.6cm}
    \label{fig:swing+foot}
\end{figure}

To demonstrate the capability of the BiconMP to generate dynamic  trajectories beyond mere gaited locomotion, we perform a high-five motion on Solo12. The goal for the robot is to give a high-five to a person in front of it by first raising both its front legs at a height above its base. Then, the robot must reach one of its arms out forward in order to high-five at a fixed position. Figure \ref{fig:hifive} shows the high-five motion generated on the robot. Since Solo12 must balance on its hind legs for the duration of the high-five, the motion needs to build significant momentum and is non-trivial to achieve. The motion planner initially gets the robot to crouch before the front two legs lift off in order to generate enough angular momentum. After the lift off, the front two legs try to reach the goal position provided by the user in the plan. The swing foot trajectory of the front left leg is shown in Fig. \ref{fig:swing+foot}. The red dots in the top three sub-plots denote the desired goal position provided by the user in the $x$, $y$ and $z$ axis. The shaded section of the plots represent the duration of the motion where the front leg is supposed to be in an air phase trying to reach the desired goal, which can be verified by noticing that the planned $z$ force becomes zero in this zone. 

Note that during this motion, the kinematics solver finds a non-trivial rotating motion for the front legs after performing the high-five in order to track the centroidal momentum trajectories provided by the biconvex centroidal dynamics solver. Specifically, the front legs of the robot swing their legs backwards and pivot around the hip joint in order to obey the momentum profile. Such a swing foot trajectory is very difficult to design a priori for a given motion which highlights the advantages of using a whole-body motion optimizer.

\begin{figure*}
\vspace{0.3cm}
\centering
\includegraphics[width=17 cm, height=6.0 cm]{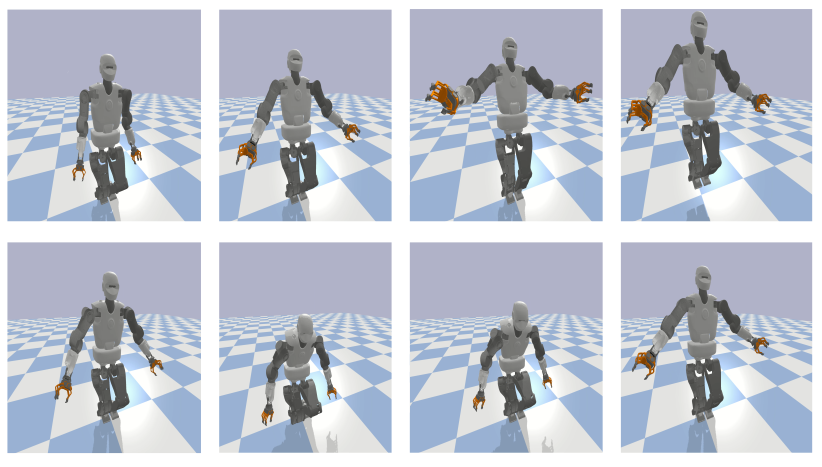}
\caption{Jump motion with Talos (Top left to bottom left in clock wise direction).}
\label{fig:talos_jump}
\end{figure*}

\subsection{Simulations with other robots}
We illustrate the generality of our approach by testing it on two other robots with different morphology, complexity and mass distributions. In all these experiments, no changes are made to the BiConMP framework (including the use the ID controller). We run these experiments in simulation. The solve times obtained for these different motions as they are run in MPC are shown in table \ref{tab:different_robots}. The resulting movements are also shown in the attached video.
\subsubsection{Talos} - The humanoid robot Talos \cite{stasse2017talos} weighs around 94 $kg$, stands at 170 cm and has 35 joints, i.e. it is larger, heavier and more complex than Solo12. Firstly, 
we validate the approach by generating stable walking in MPC. Only a reference CoM height, a Raibert controller based contact plan at $0.3s$ of step time and nominal mid swing phase via-points are provided to our optimizer. Note that in this case the MPC loop
is run at 10Hz as the mean solve time slightly exceeds 50ms. This is due to the fact that we optimize full-body motions with 35 joints now (instead of 12 for Solo12).
Secondly, we generate a jump, which is a highly dynamic motion, in MPC. The framework is only provided a contact plan which specifies a flight and stance time of $0.4$ seconds. The planner then automatically generates the required force profile to be applied on the ground and the whole-body motions that track the momentum profiles stably. This again demonstrates the strength of using a nonlinear kinematics solver to automate swing foot trajectory generation. The resulting motion in simulation is shown in figure \ref{fig:talos_jump}.  
\subsubsection{AnYmal} - The quadruped robot AnYmal \cite{hutter2016AnYmal} weighs around 30 $Kg$ which is significantly heavier than Solo12. We generate low stepping frequency trot of $0.4$ seconds and cyclic jumping with an air time of $0.3$ seconds. The solve times of AnYmal trot are slightly higher as compared to Solo12 since we use more collocation points in the trajectory. We also demonstrate stable unplanned gait transitions just like with Solo12. 
\begin{remark}
Note that we still track these motions on these robot without the need for a complicated QP based ID controller even though AnYmal and Talos' legs are heavy (i.e. contribute significant momentum) as compared to Solo12. Our framework remains unaffected even with low stepping frequencies (Anymal Trot) and arm swinging motions (Talos Jump). 
\end{remark}
\begin{table*}[h!]
\centering
\begin{tabular}{||c|c|c|c|c|c||}
\hline
Robot & Motion & Replan Frequency (Hz) & Mean Solve Time (sec) & Number Of Collocation Points & Time Discretization $\Delta t$ (sec) \\
\hline
Talos   & Jump & 10 & 0.031 &  10 &  0.04 \\
\hline
Talos   & Walk & 10 & 0.052 & 10 & 0.1 \\
\hline
AnYmal  & Trot & 20 &  0.042 & 18 & 0.05 \\
\hline
AnYmal  & Jump & 20 &  0.034 & 10 & 0.05 \\
\hline
\end{tabular}
\caption{\label{tab:different_robots} Solve times for various motions generated on different robots in simulation.}
\end{table*}

\section{Analysis of the Model Predictive Controller}
In this section, we analyse the contribution of the inverse dynamics (RNEA) controller on the stability of the robot and the advantages of using a nonlinear MPC setting on the robot as compared to pure trajectory optimization (when a fixed plan is tracked on the robot). Further, we thoroughly analyze the impact of horizon and replanning frequency on the performance and robustness of the MPC framework in order to gain insights on tuning parameters in the case of MPC for legged robotics. We also discuss qualitative advantages that were observed on the robot, which are difficult to quantify.

\subsection{Inverse Dynamics controller} 
\begin{figure}
 \vspace{0.3cm}
    \centering
    \includegraphics[scale = 0.3]{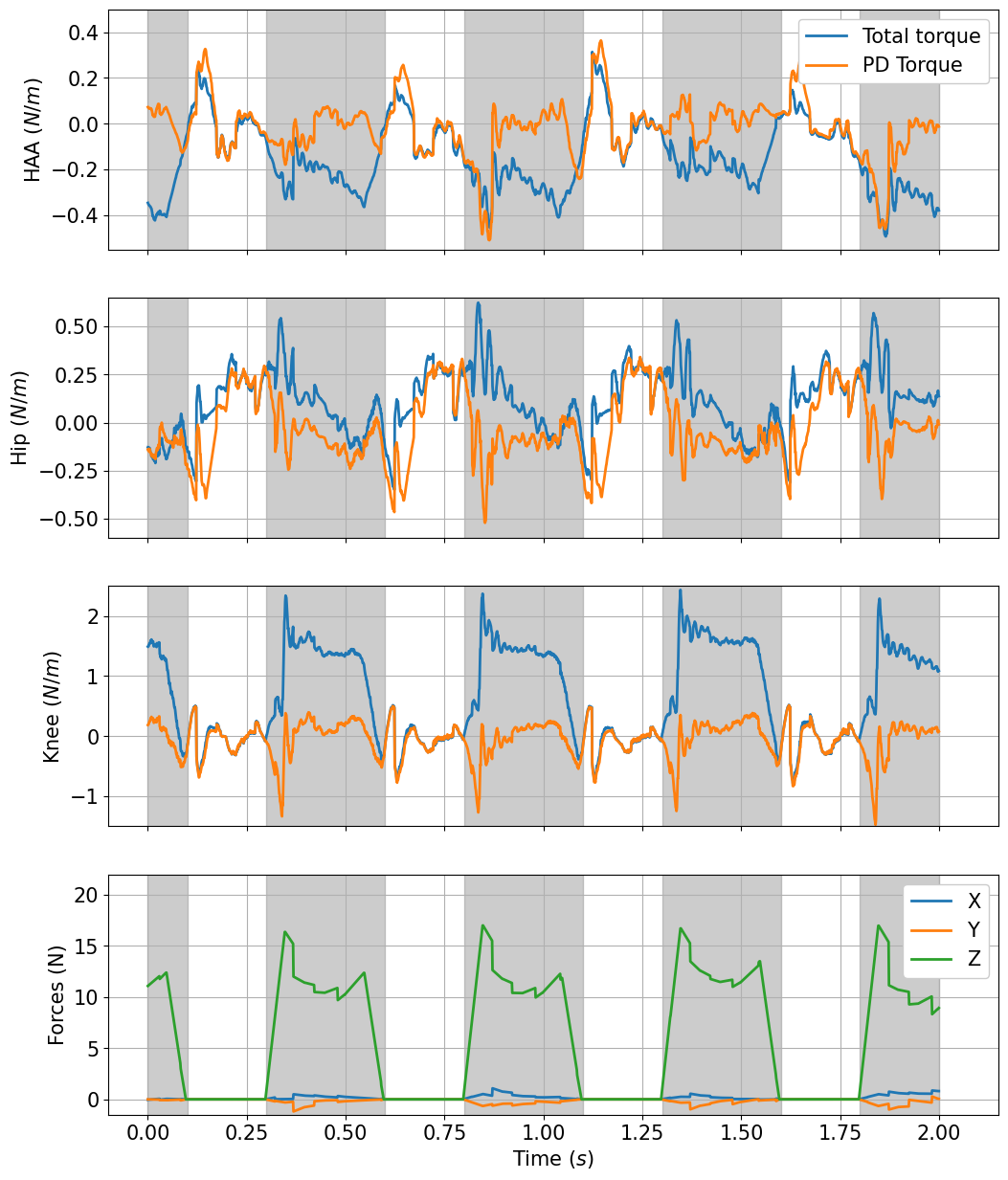}
    \caption{Contribution of PD gains to the total torques. The three joints are the hip abduction adduction (denoted HAA), the hip flexion extension (Hip) and the knee joint (Knee).}
    \vspace{-0.6cm}
    \label{fig:torques}
\end{figure}

We compare the contribution of the PD gains with the MPC planned trajectory in the inverse dynamics controller. In Fig. \ref{fig:torques}, the total torque sent to the robot along with the torque resulting from the PD controller is plotted for three joints belonging to the front left leg during the periodic jumping gait. The shaded sections highlight the contact phases of the leg during the motion. The torques due to the PD controller are very close to zero for most of the contact phase while the total torque is not zero (especially for the knee joint which does most of the work in the jump motion). This shows that the joint torques computed by the BiConMP account for most of the control during contact phases, underlining the quality of the optimized control trajectories. 
However, the PD controller seems to play a significant role during an air to contact phase transition where there is an sudden change in the planned contact forces. During run time, the optimized forces from the BiConMP are interpolated to match the low level control frequency (i.e. we have a 20Hz MPC loop and a 1KHz torque control loop). Consequently, the forces are interpolated from zero to the next desired value which also prevents sudden jumps in the commanded torques. Consequently, the motion run on the robot is slightly different from the plan during these transitions. We hypotheses that it is the reason why the contribution of the PD controller seems to be larger during these transitions. We made qualitatively similar observations for other movements and simulation experiments on the other robots (Talos and Anymal).

\subsection{Sim-to-Real Transfer} Initially, the motions discussed in section \ref{cyclic} and \ref{acyclic} were first validated in simulation (Raisim \cite{hwangbo2018per}). During this stage, the cost functions (weight tuning of the optimization problem) and gait parameters  were altered until a desirable motion was observed in simulation, after which the motion was tested on the real system. All the transfers from the simulation to the real robot was instantaneous. That is, any motion that was stable in simulation for a given set of weight parameters and gains worked directly on the robot without further tuning. This suggests that the MPC framework is sufficiently robust to model mismatch to ensure direct sim-to-real transfer. Further, it is interesting to note that the same set of $K_{p}$ gains of 3.0 and $K_{y}$ gains of 0.05 were used across all the motions presented in the result section for both simulation and real robot experiments with Solo12. In our experience, this is rarely the case when using pure trajectory optimization \cite{ponton2021efficient} without online re-planning. Often different gains are needed across different motions in simulation and these gains need to be further adapted on the robot. This is a significant benefit provided by the closed-loop MPC setup as gain tuning on the real robot is often cumbersome.  

\begin{figure}
 \vspace{0.3cm}
    \centering
    \includegraphics[scale = 0.4]{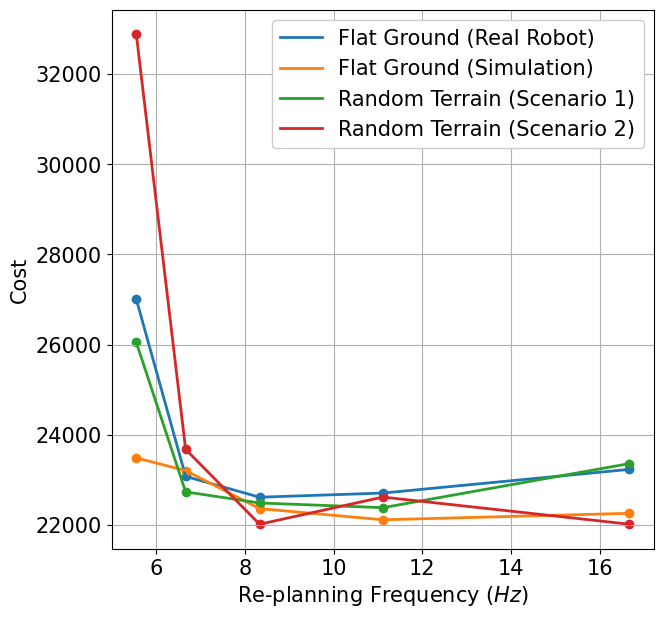}
    \caption{Comparison of MPC performance vs. replanning frequency. The change in the mean optimal cost returned by the MPC is plotted against replanning frequency.}
    \vspace{-0.6cm}
    \label{fig:cost_analysis}
\end{figure}

\subsection{Impact of re-planning frequency on performance} \label{mpcvsfreq} 
In this second set of experiments, we seek to understand the effect of the re-planning frequency on the performance of the MPC. This is an important factor, albeit seldom analyzed, as this sets minimum requirements on optimizer performances.
During this experiment, we choose the trot gait and make Solo12 track a fixed desired velocity trajectory from the same starting point. The desired velocity trajectory is set to 0.5 m/s for a fixed duration of 6 seconds after which the desired velocity is changed briefly to 0 m/s and then finally to -0.5 m/s for 6 seconds. During each run, the robot is kept at the same starting position, the weights, horizon length of the plan and the gains in the ID controller are all kept constant. Only the re-planning frequency is changed during each run. 
The mean optimal cost returned by the BiConMP is used as a metric \cite{grune2012nmpc, mayne2014model} to evaluate the performance of the MPC. We ran this experiment 4 times in which 2 times Solo12 walks on flat ground while the other two times it walks on random terrain (as shown in Fig. \ref{fig:motions}). The same experiment is also run on flat ground in simulation for further comparison. 

The results obtained from the experiments are shown in Fig. \ref{fig:cost_analysis}. The plot shows that after a certain threshold re-planning frequency is reached the MPC performance does not change. In this case the threshold re-planning frequency is approximately 7 Hz. Further, this threshold does not change even in the presence of terrain uncertainties. Consequently, this suggests that after reaching a desired threshold there does not seem to be any benefit, in re-planning much faster with respect to performance.  Further, the performance of the MPC in simulation as compared to the real robot is also very similar after this threshold re-planning frequency is reached. This result could also help explain why we were able to the directly transfer motions from simulation to the real robot without any additional changes (i.e. our re-planning frequency was sufficiently high).

\begin{figure}
 \vspace{0.3cm}
    \centering
    \includegraphics[scale = 0.5]{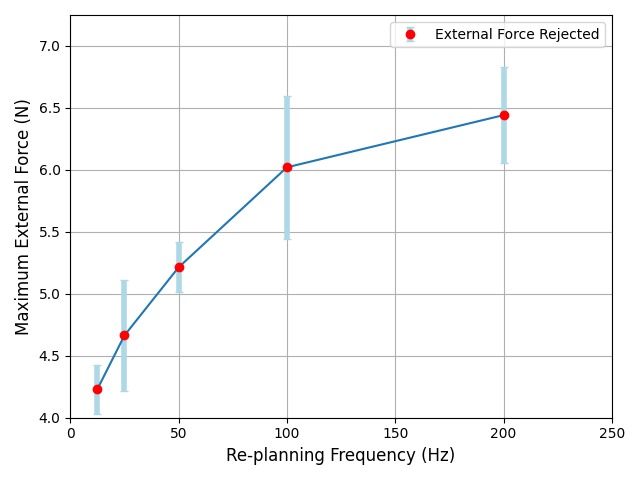}
    \caption{Comparison of maximum external disturbance vs. re-planning frequency for 3 different motions: trotting, bounding, and jumping. Experiment performed in simulation. }
    \vspace{-0.6cm}
    \label{fig:maximum_disturbance}
\end{figure}

\subsection{Impact of planning horizon on performance} 
Another important parameter in MPC is the length of the optimization horizon. To reduce computational cost, it needs to be as small as possible but a too short horizon (depending on the choice of terminal cost) will lead to unstable behaviors.
In this third set of experiments, we analyse how the length of the planning horizon influences the performance of the MPC. The experimental setup was kept identical to the previous subsection (\ref{mpcvsfreq}). The only difference here is that the planning horizon is changed while the re-planning frequency is kept constant. For these experiments, we were unable to determine a good evaluation metric. The cost function can not be used as a metric since the total value attainable by the cost changes with problem size (number of collocation points) which in turn depends on the horizon. Consequently, we only discuss qualitative results observed during the experiments. 

For a low re-planning frequency where a stable gait is not observed (frequency lower than 7 Hz, section \ref{mpcvsfreq}), we observed that performance/stability could be improved by increasing the length of the planning horizon. After a threshold length is reached, additional increases did not seem to bring any visible benefits. In the presence of terrain uncertainty, a similar result was observed. Further, the threshold planning horizon for a given re-planning frequency remained the same with or without terrain uncertainty. Consequently, we found that at low re-planning frequency, the stability of the motion could be increased by increasing the horizon. This result is in alignment with well-established theoretical results \cite{mayne2014model}.

\subsection{Impact of re-planning frequency on robustness} In the final set of experiments, we analyse how overall motion robustness increases as the re-planning frequency increases. 
Here we choose to specifically evaluate robustness to external disturbances. 
Since measuring during locomotion, in a reproducible manner, the maximum magnitude of disturbance rejection is difficult on the real system, we perform this experiment in simulation. Moreover, we test the robustness at very high frequencies which are not realizable in the real world (solve times are higher than the re-planning time). However, our previous experiments suggest that results from the simulation should carry to the real robot.
In each run, an external force is applied on the base in a random direction discretized at intervals between 0 and 360 degrees. We push the robot for a duration of 0.2 to 0.5 seconds. We run each experiment 10 times for each duration and run the experiments on three different gaits (trotting, bounding and jumping.

The results are shown in Fig. \ref{fig:maximum_disturbance}. This experiment shows that there is a gain in robustness, in terms of disturbance rejection when moving from 20 Hz to 100 Hz, after which the relative gain starts to decay. Based on the previous experiment (Sim-to-Real and MPC performance), a similar result could be expected on the real system provided the solver was 4-5 times faster as the solution quality of the MPC is almost the same in simulation and real robot beyond a re-planning frequency of at least 10 Hz.

\section{Discussion}
This section discusses the proposed approach
and the experimental results with respect to the current state of the art.

\subsubsection{Algorithms for closed-loop whole-Body MPC}
Very few algorithms have demonstrated closed-loop whole-body MPC on real legged robots. General purpose interior-point or sequential quadratic programming (SQP) methods are not capable of providing solve times low enough for real-time control.
Most of the existing approaches use custom DDP-like methods to solve the entire whole-body optimization problem at once \cite{neunert2018whole, dantec2021whole} or decompose it \cite{sleiman2021unified}. These approaches have demonstrated closed-loop MPC ranging from $20-80$ Hz for particular motions like trot and jumps. However, these algorithms have seldom been used to show diverse and dynamic motions on robots with different anatomy in closed loop.
To our knowledge, our method is the first algorithm using ADMM and first-order proximal methods to demonstrate closed-loop MPC at competitive rates.

The advantage of using DDP-like methods is that they can in principle solve any OCP and efficiently exploit the time-induced sparsity in the problem. DDP further provides optimal feedback gains that can be used for local high-frequency control. However, enforcing constraints becomes challenging (typically hard constraints are not enforced in reported MPC results). 
The most common practice of using log-barriers to enforce constraints can be numerically problematic \cite{grandia2019feedback} due to ill-conditioning of the Hessian. To address this issue, constraints can be relaxed in order to find reasonable solutions \cite{grandia2019feedback}. Recently, DDP-based algorithms that enforce constraints directly with low solve times have been proposed \cite{jallet2022implicit}. However, 
they are yet to be demonstrated on high-dimensional problems in MPC.\\
On the other hand, the BiConMP does not solve general optimal control problems and largely relies on the structure
of the floating-based dynamics (kino-dynamic decomposition and biconvex structure of the centroidal dynamics). This has the advantage that constraint enforcement, especially on contact forces, becomes rather straightforward due to the use of the proximal operators (including second-order cones) without any loss in the solve times or convergence rates.
Furthermore, problems related to Hessian ill-conditioning inherent to second order methods are removed with the use of FISTA (a first order method).
While we exploited the structure of the centroidal dynamic,
we still use DDP to solve the kinematic optimization problem. It might be interesting
to explore whether this problem can be further decomposed to exploit proximal methods and potentially improve efficiency while also including hard constraints.
Force constraint enforcement plays an important
role (as already highlighted in \cite{Righetti2013, Viereck2022})
and our results suggest that this helps find control trajectories that can be directly executed on the robot with a simple inverse dynamics controller, without the need for an additional dynamic filter (as a QP-based inverse dynamics) unlike DDP-based approaches \cite{sleiman2021unified}.

In general, existing DDP-based approaches need a good warm start trajectory \cite{dantec2021whole, lembono2020learning, neunert2018whole} to actually achieve high replanning rates. Indeed, quadratic convergence rates are only guaranteed close to a local minimum and DDP can be quite slow away from it.
Obtaining these trajectories is, however, a challenge which still limits the applicability of DDP-based whole-body solvers. 
In contrast, we did not encounter the need for good initialization in any of our experiments. The ability of ADMM to converge to good solutions quickly allows early termination of the solver. This property is favourable in closed loop MPC settings with real-time requirements and is not present with other second order methods.
This property could also be exploited other problems with closed loop MPC settings.


\subsubsection{Advantages of proximal methods for MPC}
The use of first order optimization methods for MPC is not common in robotics, nor
is the use of general proximal methods. We believe that beyond the legged robots, 
our work highlights a few interesting properties more
broadly applicable.
First, algorithms such as FISTA are very easy to implement (they only need gradients) and are numerically robust. 
Since sensor noise can limit the availability of true gradients, first order methods are more likely to converge to good solutions compared to second-order methods \cite{mohammadi2020robustness,liang2022improving}. Indeed, we observed that the algorithm was surprisingly stable in a closed loop MPC setting. Further, the use of proximal operators renders the constraint satisfaction problem rather easy.
This suggests that first order methods might play an increasing role for MPC solvers in robotics. Furthermore, developing custom solvers based on proximal operators and related augmented Lagrangian formulations \cite{jallet2022implicit, jackson2021altro} for other closed loop application is also a promising research direction, especially when early termination is more important than very high precision.

\subsubsection{Enforcing torque constraints}
The main assumption in the kino-dynamic decomposition of the nonlinear robot dynamics is that there exists sufficient torque authority \cite{herzog2016structured}, without which computing feasible torques becomes impossible (or needs several kino-dynamic iterations) for the given plan. During our experiments on the robot, this assumption has never been violated for all the motions even though only one dynamic to kinematic iteration is performed. Further, the computed torques have been much lower than the maximum torque limits of Solo12 and subsequently more aggressive behaviours can be performed if needed. In the case that this limit is being reached, more than one kino-dynamic iteration can be performed to ensure better consensus. It is also possible to add torque constraints in the kinematic optimizer at the cost of slightly higher solve times as discussed in previous work \cite{ponton2021efficient}.

\subsubsection{Insights from Nonlinear MPC Implementation}
Running the nonlinear whole body MPC has shown several advantages on the robot along with a few key insights: re-planning online in general improves the robustness of the robot to disturbances and terrain. In addition, the whole body optimization allows the robot to automatically change swing foot trajectories without highly specified references. An interesting result from our analysis is that increasing the re-planning frequency or horizon above a certain threshold does not seem to give any major advantages in terms of performance for the tasks we analyzed. However, there is a significant improvement in robustness as the frequency is increased from 20 Hz to 100 Hz after which the rate of gain starts to decrease. Consequently, this analysis suggests that re-planning frequencies higher than 10 Hz are not needed to achieve direct sim to real transfer for the Solo12. However, to gain more robustness to external disturbances, higher frequencies are needed and avenues such as warm starting the solver, optimizing the implementation or further exploiting the problem structure can be explored.

\subsubsection{Comparison to Deep Reinforcement Learning (DRL)}
Recently, DRL has become an increasingly popular choice to generate robust trajectories for legged robots \cite{lee2020learning,siekmann2021blind}. 
One main reason stems from the fact that 
MPC approaches need fast optimizers while DRL approaches learn a policy offline which is rather cheap to evaluate online.
However, our proposed method has a re-planning frequency comparable to these methods, with the scope of further speed up in future work. In addition, generating new trajectories with an optimizer is significantly less cumbersome and does not require one to re-train a policy for different types of motions. In our case, the same cost function with different weights can be used to generate different motions. Finally, the sim-2-real transfer is simple and instantaneous in our approach as the BiConMP is able to compensate for modeling errors automatically thanks to the closed-loop optimization. On the other hand sim-2-real transfer with DRL methods is usually not simple because they depend heavily on the trained robot model.  Subsequently, they require very accurate robot actuator models \cite{hwangbo2018per} in simulation or domain randomization is necessary \cite{siekmann2021blind} for successful transfer. 


\section*{Conclusion}
We proposed a nonlinear MPC framework, the BiConMP, capable of generating dynamic behaviors in real-time for various legged robots. We exploit the biconvex nature of the centroidal dynamics to
propose an efficient solver based on ADMM and proximal gradient methods. We further propose 
to formulate the kinematic problem as an optimal control problem which is then solved using off-the-shelf DDP solvers \cite{mastalli2020crocoddyl}. 
Through various real robot and simulation experiments
we demonstrated the ability of the approach to generate and control very dynamic movements.
We conducted an extensive analysis of the various parameters of the MPC framework such as frequency, cost, and horizon to understand their impact on the performance and robustness on the real robot, hence suggesting general guidelines for MPC requirements.
In future work, we intend to investigate the effect
of warm-start on the solver efficiency. We also
intend to further explore the capabilities of first
order proximal methods for more general MPC applications to robotics.

%

%

\bibliography{master}
\bibliographystyle{IEEEtran}

\end{document}